\documentclass[10pt,journal,compsoc]{IEEEtran}

\ifCLASSOPTIONcompsoc
  \usepackage[nocompress]{cite}
\else
  \usepackage{cite}
\fi

\usepackage{graphicx} 
\usepackage{algorithm}
\usepackage{algpseudocode}
\usepackage{amsmath}
\usepackage{bbm}
\usepackage{algorithm}
\usepackage{algpseudocode}
\usepackage{color}
\usepackage{hyperref}
\usepackage[usenames,dvipsnames]{xcolor}
\usepackage{tikz}
\usetikzlibrary{shapes.geometric, arrows}
\usetikzlibrary{calc}

\ifCLASSINFOpdf
\else
\fi

\begin{document}
\title{Evolutionary Clustering via Message Passing}

\author{Natalia~M.~Arzeno and~Haris~Vikalo,~\IEEEmembership{Senior Member,~IEEE}
\IEEEcompsocitemizethanks{\IEEEcompsocthanksitem N. M. Arzeno was with the Department
of Electrical and Computer Engineering, The University of Texas at Austin. \protect
E-mail: narzeno@gmail.com.
\IEEEcompsocthanksitem H. Vikalo is with the Department of Electrical and Computer Engineering,
The University of Texas at Austin. \protect E-mail: hvikalo@ece.utexas.edu.
}
}

\markboth{Evolutionary Clustering via Message Passing}%
{Shell \MakeLowercase{\textit{et al.}}: Bare Demo of IEEEtran.cls for Computer Society Journals}
\IEEEtitleabstractindextext{%
\begin{abstract} We are often interested in clustering objects that evolve over
time and identifying solutions to the clustering problem for every time step. Evolutionary clustering provides insight into cluster evolution and temporal changes in cluster memberships while enabling performance superior to that achieved by independently clustering data collected at different time 
points. In this paper we introduce evolutionary affinity propagation (EAP), an evolutionary clustering algorithm 
that groups data points by exchanging messages on a factor graph. EAP promotes temporal smoothness of
the solution to clustering time-evolving data by linking the nodes of the factor graph that are associated with 
adjacent data snapshots, and introduces consensus nodes to enable cluster tracking and identification 
of cluster births and deaths. Unlike existing evolutionary clustering methods that require additional processing 
to approximate the number of clusters or match them across time, EAP determines the number of clusters and 
tracks them automatically. A comparison with existing methods on simulated and experimental data demonstrates 
effectiveness of the proposed EAP algorithm.

\end{abstract}

}

\maketitle
\IEEEdisplaynontitleabstractindextext
\IEEEpeerreviewmaketitle

\let\thefootnote\relax\footnote{\normalfont \copyright 2019 IEEE.  Personal use of this material is permitted.  Permission from IEEE must be obtained for all other uses, in any current or future media, including reprinting/republishing this material for advertising or promotional purposes, creating new collective works, for resale or redistribution to servers or lists, or reuse of any copyrighted component of this work in other works.}

\IEEEraisesectionheading{\section{Introduction}\label{sec:introduction}}

\IEEEPARstart{I}n a number of applications we are interested in clustering data whose features evolve over time. Examples include identification of communities in dynamic social networks \cite{tant07}, tracking objects \cite{Li:2004aa}, and analysis of time-series financial data \cite{Fenn:2009aa}. Heuristic approaches to learning structure of temporally evolving data typically perform cluster analysis for each snapshot independently and then attempt to relate clustering solutions across time. However, such methods are often very sensitive to noise and short-term perturbations, and generally struggle with cluster tracking. An alternative to independent analysis of data snapshots is to perform evolutionary clustering, i.e., seek to organize data collected at multiple points in time while taking into account underlying dynamics and promoting temporal smoothness of the resulting clusters. Such an approach is found to typically be more informative and generally outperforms 
clustering conducted independently at each time point \cite{chakrabarti_2006, ahmed_2008, xu_2013}. 
In recent years, traditional clustering algorithms such as k-means, spectral clustering, and agglomerative clustering have been adapted to the evolutionary clustering setting 
\cite{chakrabarti_2006,chi_2007,chi_2009}, and used in a wide range of applications 
\cite{chakrabarti_2006, chi_2009, xu_2010,czink_2007}. These evolutionary 
clustering algorithms modify the objective of traditional clustering problems to include both a term that 
quantifies quality of the clustering results at each time step as well as a temporal smoothness term that promotes 
sustained cluster membership.

Inferring the number of clusters, often done heuristically, is a major challenge for evolutionary as well as traditional clustering methods. Existing evolutionary clustering methods that 
attempt to automatically decide the number of clusters typically rely on non-parametric Bayesian techniques 
(specifically, Dirichlet process models \cite{ahmed_2008, ahmed_2010, xu_2008}). Ideally, evolutionary 
clustering algorithms should be capable of detecting changes in the number of clusters as data evolves, 
i.e., they should allow 
clusters to be born, evolve, or die at each time step. Moreover, they should be capable of handling data 
points that appear or disappear over time. While there exist algorithms that can satisfy some of these 
requirements \cite{chi_2009,xu_2013,ahmed_2008,folino_2014}, practically feasible solutions to the 
evolutionary clustering problem remain elusive.

In this paper we propose evolutionary affinity propagation (EAP), a clustering algorithm that builds upon 
ideas of static affinity propagation (AP) to organize data acquired at multiple points in time by passing 
messages on a factor graph; specifically, the graph allows exchange of 
information between nodes associated with different time-snapshots of the data. EAP automatically 
determines the number 
of clusters for each time step and, similar to AP, can handle non-metric similarities and efficiently process 
large sparse datasets. In a departure from AP, EAP relies on consensus nodes that we introduce to accurately track clusters across time and identify points changing cluster membership.
Moreover, EAP can detect cluster births and deaths as well as handle data insertions and 
deletions. Designed to search for the global clustering solution, EAP avoids error propagation 
that adversely affects performance of existing evolutionary clustering methods; unlike EAP, those methods 
form a solution at time $t$ using only the data (or the clustering solution) at $t-1$ while disregarding 
data at other times. Note that EAP takes a data-centric approach to clustering and tracks individual 
points across time. This stands apart from the distribution-based evolutionary clustering methods 
which focus exclusively on data generative distributions and attempt to infer evolution of the parameters 
of those distributions \cite{ahmed_2008, ahmed_2010, xu_2008}; for this reason, distribution-based 
methods typically require an additional cluster assignment step. To our knowledge, EAP is the first 
evolutionary clustering algorithm that automatically detects the number of clusters, automatically 
tracks clusters across time, and focuses on data instead of distribution models. 


The paper is organized as follows. In Section \ref{sec:background}, existing evolutionary clustering 
methods and the traditional affinity propagation algorithm are overviewed. The EAP algorithm is presented in Section \ref{sec:methods} and compared to existing schemes in Section \ref{sec:results}. Potential future directions are outlined in
Section~\ref{sec:future} and the paper is concluded in Section \ref{sec:conclusions}. Our preliminary
 work that introduced the basic ideas of linking variable nodes of a factor graph across time and 
 creation of consensus nodes was reported in conference paper \cite{arze17}. The current paper goes beyond \cite{arze17} by providing derivation of EAP messages; presenting algorithms for exemplar identification, tracking cluster evolution, and detection of cluster deaths; providing a detailed study of the effects of parameter values on the performance of EAP; proposing methodology for handling insertion and deletion of time points; and presenting extensive benchmarking tests of EAP on real-world oceanographic, financial and healthcare data.

\section{Background} \label{sec:background}

\subsection{Evolutionary clustering}
 
Chakrabarti et al.'s landmark paper \cite{chakrabarti_2006} introduced evolutionary k-means and evolutionary 
agglomerative hierarchical clustering; the approach proposed there promotes temporal smoothness of
clustering solutions by optimizing an objective that at a given time consists of a snapshot quality term 
and a historical cost term. Evolutionary spectral clustering followed soon thereafter \cite{chi_2009}, adopting the 
same framework while allowing one to choose whether to place more emphasis on preserving cluster quality 
or cluster membership. In \cite{xu_2013}, Xu et al. proposed AFFECT, an evolutionary clustering method that 
represents the matrix indicating similarity between data points at a given time as the sum of a deterministic 
proximity matrix and a Gaussian noise matrix. The AFFECT framework enables adaptation of the classic k-means, 
agglomerative, and spectral clustering algorithms to evolutionary setting, while allowing optimization of the weight 
of temporal cost term in the objective function. A non-parametric Bayesian approach to evolutionary clustering 
that relies on Dirichlet process models to discover the number of clusters was studied in 
\cite{ahmed_2008,ahmed_2010,xu_2008}. This line of work includes a scheme where cluster parameters evolve 
in a Markovian fashion and the posterior optimal cluster evolution is inferred by Gibbs sampling 
\cite{ahmed_2008,ahmed_2010}, and a method with an automatic cluster number inference that combines a 
hierarchical Dirichlet process with a transition matrix from an infinite hierarchical hidden Markov model 
\cite{xu_2008}. Note that the aforementioned algorithms require some form of post-processing in order to
match clusters across different time steps and enable tracking of cluster dynamics including cluster evolution, 
appearance of new clusters, and cluster dissolution; for an illustration of the required post-processing steps 
see, e.g., \cite{greene_2010, brodka_2012}.

In recent years, evolutionary clustering has been applied in a variety of settings including climate change studies 
\cite{gunnemann_2011}, analysis of categorical data streams \cite{kim_2015}, and detection and tracking of 
web user communities \cite{yang_2010,fortunato_2010, folino_2014, jia_2014}. Note that even though community 
detection is not explored in the current paper, EAP can be used in that setting assuming a feature vector is available 
(e.g., users' contribution to different types of web forums or the number of people in different categories followed 
on Twitter).



\subsection{Affinity propagation}

\begin{figure}[h]
\vspace{-0.1in}
\caption{A subgraph of the factor graph showing subset of variable and factor nodes associated with points 
$i$, $j$, $k$ and $l$.} \label{APgraph}
\includegraphics[width=2.5in]{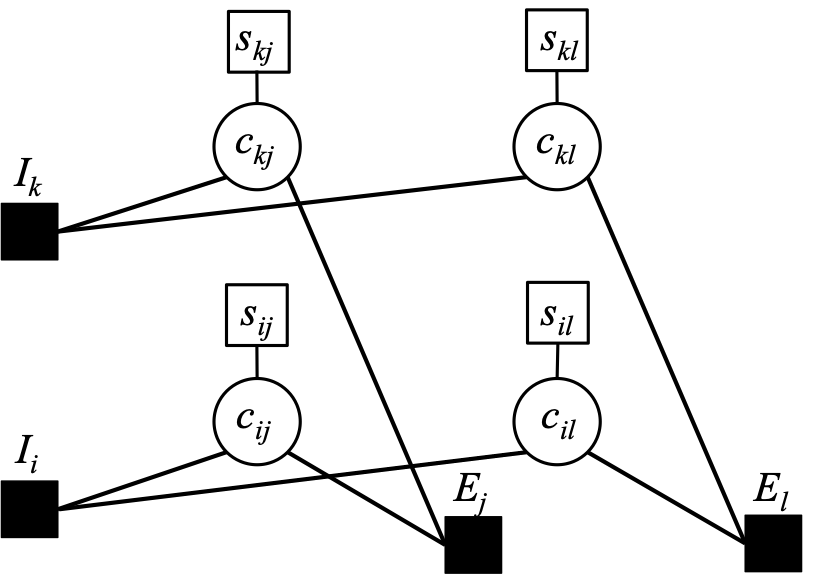}
\centering
\end{figure}

Affinity propagation (AP) is a clustering algorithm that seeks to group data points by exchanging messages 
between nodes of a graph representing the data \cite{frey_2007}. For each cluster, the algorithm identifies
an ``exemplar" -- a member of the data set that represents points in the cluster. The resulting 
clusters may be interpreted as subgraphs spanned by edges that connect points with their exemplars. 
Similarity between the nodes in the graph is specified using a measure such as the negative Euclidean 
distance or the Pearson correlation coefficient. The objective of AP is to maximize the total similarity 
between points and their exemplars,
\begin{equation} \label{APoptproblem}
\max_{c_{ij}} \sum_{i=1}^{N}\sum_{j=1}^{N}s_{ij}c_{ij}+\sum_{i=1}^{N}I_i(\{c_{ik}\})+\sum_{j=1}^{N}E_j(\{c_{kj}\}),
\end{equation}
where $s_{ij}$ denotes similarity between points $i$ and $j$, binary variable $c_{ij}$ indicates if $j$ is an exemplar 
of $i$, while
\begin{align*}
I_i(c_{i1},c_{i2}\ldots c_{iN})&=\begin{cases}-\infty & \text{if~} \sum_{j=1}^{N}c_{ij} \neq 1, \\ 0 & \text{otherwise,} \end{cases}
\end{align*}
\begin{align*}
E_j(c_{1j},c_{2j}\ldots c_{Nj})&=\begin{cases}-\infty & \text{if~} c_{ij}=1, c_{jj} \neq 1, \\ 0 & \text{otherwise,} \end{cases}
\end{align*}
enforce single-cluster membership and exemplar self-selection constraints, respectively. To solve
(\ref{APoptproblem}), AP relies on exchanging messages between nodes of a factor graph having variable
nodes associated with $c_{ij}$ and factor nodes associated with $I_i(c_{i1},c_{i2}\ldots c_{iN})$ and 
$E_j(c_{1j},c_{2j}\ldots c_{Nj})$. A subgraph showing variable and factor nodes linked with points 
$i$, $j$, $k$ and $l$ is shown in Fig.~\ref{APgraph}.
AP requires exchange of only two messages, responsibility and availability, between data points. The 
responsibility $\rho_{ij}$ indicates suitability of point $j$ to be an exemplar for point $i$ while the availability 
$\alpha_{ij}$ expresses the level of confidence that point $i$ should choose $j$ as an exemplar. These 
messages are derived from the max-sum algorithm on the aforementioned factor graph \cite{givoni_2009}, 
resulting in $\rho_{ij} =s_{ij}-\max_{k \neq j} (\alpha_{ik}+s_{ik})$ and
\begin{align*}
\alpha_{ij}=& \begin{cases}
\sum_{k\neq j}\max[\rho_{kj},0] & \text{if } i=j, \\
\min \big{[} 0,\rho_{jj}+\sum_{k\neq \{i,j\}}\max[\rho_{kj},0]\big{]} & \text{if }i\neq j.
\end{cases}
\end{align*}
To avoid numerical oscillations, messages are often damped; for instance, the update for $\rho_{ij}$ in iteration $k$ is calculated as
\begin{equation}
 \rho_{ij}^{(k)} =\lambda\rho_{ij}^{(k-1)} + (1-\lambda) \big(s_{ij}-\max_{k \neq j} (\alpha_{ik}+s_{ik})\big),
\label{eq_damping}
\end{equation}
where $\lambda$ is the damping factor. The damping is applied in a similar way to the 
$\alpha_{ij}$ message updates. AP does not require similarities to be metric \cite{frey_2007}, 
and can be efficiently implemented to cluster large, sparse datasets by only passing messages 
between points that have a similarity measure. The number of clusters is automatically inferred
by the algorithm and can be tuned via self-similarity, or preference, of the data points if such
prior information is available.

A number of extensions of AP have been proposed in recent years, including semi-supervised 
clustering with strict \cite{givoni_2009ss,leone_2008} or soft \cite{arzeno_2015} pairwise 
constraints, relaxation of the self-selection constraint \cite{leone_2008}, hierarchical AP 
\cite{givoni_2012}, AP with subclass identification \cite{wang_2013}, and fast AP with adaptive 
message updates \cite{fujiwara_2015}. AP has further been applied to studies of the dynamics of shoals (groups of fish traveling together) \cite{quera_2013}, 
where an algorithm referred to as soft temporal constraint affinity propagation employs modified 
availability messages to impose preference of assigning points at time $t+1$ to the same 
exemplar as at time $t$. However, this scheme does not impose backward temporal smoothness, 
would require additional post-processing steps to attempt tracking clusters, and would struggle if the number of objects varies between time steps as objects emerge or disappear.

\section{Methods: Evolutionary AP}\label{sec:methods}

In this section, we first present a factor graph that enables exchange of responsibility and availability 
messages over time and 
thus facilitates evolutionary affinity propagation. We then expand on this basic framework by 
introducing consensus nodes which allow for more accurate tracking of the clusters and enable
detection of cluster births, monitoring their evolution, and inference of their death.

\subsection{Basic evolutionary affinity propagation framework}
\label{sec:basicmethods}

\begin{figure}[h] 
\caption{Factor graph for evolutionary affinity propagation.} \label{figure_EAP}
\includegraphics[width=3.25in]{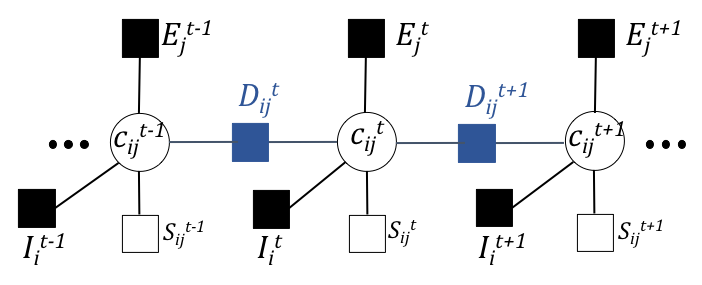}
\centering
\vspace{-0.1in}
\end{figure}

Recall that the traditional AP algorithm clusters data points collected at each time step independently of other 
time steps; unfortunately, this may not reveal the true structure of the temporally evolving data and generally 
leads to different exemplars for the same cluster at different times. To address this problem, EAP clusters data 
by exchanging messages on the factor graph shown in Fig.~\ref{figure_EAP}; this structure consists of 
subgraphs associated with individual time steps that are linked by novel factor nodes $D_{ij}^t$. In particular, 
$D_{ij}^t$ establish connection between the variable nodes of the subgraphs and render responsibility messages 
dependent on messages from previous and subsequent time steps. Exchanging messages across time enables 
us to penalize clustering configurations where data points repeatedly change exemplars and thus helps promote 
temporal smoothness of the solution to the clustering problem. The final configuration of clusters (i.e., groupings 
of points in each of the time steps) is the result of considering all points as potential exemplars at all time steps 
while encouraging exemplar stability and temporal smoothness. 
Including exemplar stability as an explicit term in the presented EAP formulation allows us to track the evolution 
of clusters without needing an additional step of matching clusters across time as, e.g., in \cite{xu_2013}. The 
cluster matching, avoided in the EAP setting, requires polynomial-time computational complexity for one-to-one 
matching and becomes more complex in general.

\begin{figure}[h]
\vspace{-0.1in}
\caption{EAP messages at time $t$.} \label{figure_EAPmessages}
\includegraphics[width=2in]{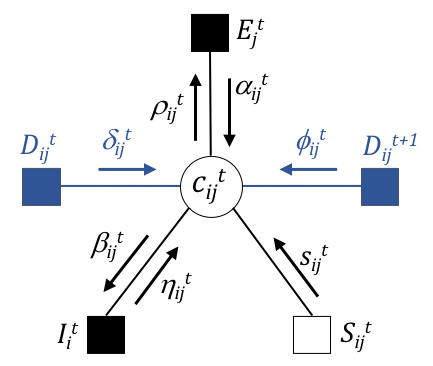}
\centering
\end{figure}

Messages exchanged between the nodes of the EAP factor graph are specified next. Note that the factor nodes 
and messages present in EAP but absent from the AP algorithm are highlighted in Fig.~\ref{figure_EAPmessages} 
in blue. As in the classical AP \cite{givoni_2009}, variable $c_{ij}^t$ in Fig.~\ref{figure_EAPmessages} takes on 
value 1 if $j$ is the exemplar for $i$ and is $0$ otherwise. Factor node $I_i^t$ ensures that each data point is 
assigned to only one cluster, $E_j^t$ enforces the constraint that if $j$ is an exemplar for any $i\neq j$ then $j$ 
must also be an exemplar for itself, and $S_{ij}^t$ passes the similarity between a point and its exemplar (i.e., 
communicates $s_{ij}^t$). Recall that node $D_{ij}^t$ encourages temporal smoothness 
by penalizing (but not prohibiting) changes in clusters' structure. Unlike in the traditional AP algorithm, values 
of the nodes in the EAP graph are time-dependent and can formally be stated as 
\begin{align}
E_j^t(c_{1j}^t,\ldots,c_{Nj}^t)=&
	\begin{cases}
	-\infty & \text{if } c_{jj}^t=0 \text{ and} \sum_i c_{ij}^t>0 \\
	0 & \text{otherwise}
	\end{cases} \nonumber \\
I_i^t(c_{i1}^t,\ldots,c_{iN}^t)=&
	\begin{cases}
	-\infty & \text{if } \sum_j c_{ij}^t\neq 1 \\
	0 & \text{otherwise}
	\end{cases} \nonumber \\
S_{ij}^t(c_{ij}^t) = & \begin{cases}
	s^t_{ij}  & \text{if } c_{ij}^t= 1 \\
	0 & \text{otherwise}
	\end{cases} \nonumber \\
D_{ij}^t(c_{ij}^{t-1},c_{ij}^t) = &\begin{cases}
	-\gamma & \!\text{if } c_{ij}^{t-1}\neq c_{ij}^t \\
	0 & \!\text{otherwise,}
	\end{cases} \label{eq_D}
\end{align}
where $\gamma > 0$.

We derive the EAP messages by relying on the max-sum update rules \cite{bishop_2006}, with bidirectional messages between variable nodes and factor nodes. Specifically, message
from a variable node to a factor node ($m_{x \rightarrow f}$) is defined as the sum of the messages arriving to the 
variable node from all other factor nodes,
\begin{equation} \label{eq_vartofac}
 m_{x \rightarrow f}(x) = \sum\limits_{g: g \in \text{ne}(x) \setminus f} m_{g \rightarrow x}(x),
\end{equation}
where $\text{ne}(x)$ denotes the neighborhood of the variable node $x$. Following \cite{givoni_2009}, 
we do not propagate a distinct message for each state of the binary variable $c_{ij}$; instead, we propagate the message difference $m_{ij}^t$ formed (for any type) as
\begin{equation} \label{eq_defmsg}
m_{ij}^t = m_{ij}^t(c_{ij}^t=1)-m_{ij}^t(c_{ij}^t=0).
\end{equation}

To more easily relate to AP, the message derivation follows a similar process. In the end, messages $\rho$ and $\alpha$ continue to be the responsibility and availability with some modifications to account for the temporal component, messages $\beta$ and $\eta$ are replaced in the derivation, and $\delta$ and $\phi$ are new messages specific to EAP.

Using the update rule for messages from a variable node to a factor node \eqref{eq_vartofac} and the definition
\eqref{eq_defmsg}, we readily derive $\beta_{ij}^t$ and $\rho_{ij}^t$ (please see Fig.~\ref{figure_EAPmessages}) 
as
\begin{align}
\label{eq_beta}
\beta_{ij}^t=&\alpha_{ij}^t+s_{ij}^t+\phi_{ij}^t+\delta_{ij}^t, \\
\rho_{ij}^t = & s_{ij}^t+\eta_{ij}^t+\phi_{ij}^t+\delta_{ij}^t. \label{eq_rho}
\end{align}
Note that, as suggested by the lack of explicit messages from $c_{ij}^t$ to $D_{ij}$ in Fig.~\ref{figure_EAPmessages}, messages from a variable node to the factor node 
$D_{ij}$ are readily determined from the sum of all other messages going into $c_{ij}^t$. 

A message sent from a factor node to a variable node is formed by maximizing the sum of messages from other factor nodes to the variable node and the current function value at the factor node,
\begin{equation}
 m_{\!f \rightarrow x}(x) \!=\!\!\!\! \max_{x_1,\ldots,x_n} \!\left[ f(x,x_1,\ldots,x_n)\!+\!\!\!\!\!\!\!\!\!\sum\limits_{k: k \in \text{ne}(f) \setminus x} 
\!\!\!\!\!\!\!\!m_{x_k \rightarrow f}(x_k) \right]\!\!.
\end{equation}

Following these definitions, $\eta_{ij}^t$ and $\alpha_{ij}^t$ remain the same\footnote{Message $\alpha_{ij}$ from 
the factor node $E_j$ to the variable node $c_{ij}$ is dependent only on messages received at $E_j$ from other 
variable nodes and not on any messages coming from the new factor nodes in EAP. Likewise, $\eta_{ij}$ is not 
affected by the new messages in EAP since it only depends on messages $I_i$ receives from other variable 
nodes.} as the corresponding messages in the traditional AP at a given time (please see \cite{givoni_2009} for the 
derivation),
\begin{align} 
\eta_{ij}^t =& -\max_{k\neq j} \beta_{ik}^t \label{eq_eta} \\
\alpha_{ij}^t=& \begin{cases}
\sum_{k\neq j}\max[\rho_{kj}^t,0] & \text{if } i=j \\
\min \big{[} 0,\rho_{jj}+\sum_{k\not\in \{i,j\}}\max[\rho_{kj},0]\big{]} & \text{if }i\neq j. \label{eq_alpha}
\end{cases}
\end{align}

Using (\ref{eq_beta}) and (\ref{eq_eta}), $\eta_{ij}^t$ can be eliminated from the definition of $\rho_{ij}^t$. In 
particular, although the updates for $\delta_{ij}$ and $\phi_{ij}$ depend on $\eta_{ij}$, we can leverage 
(\ref{eq_rho}) to substitute $\eta_{ij}^t =\rho_{ij}^t-s_{ij}^t-\phi_{ij}^t-\delta_{ij}^t$ and remove message 
dependence on $\eta$. The responsibilities $\rho$ can then be rewritten as
\begin{equation}
\rho_{ij}^t=s_{ij}^t+\phi_{ij}^t + \delta_{ij}^t -\max_{k\neq j}(\alpha_{ik}^t+s_{ik}^t+\phi_{ik}^t + \delta_{ik}^t).
\end{equation}

Therefore, there are four messages that need to be computed for each pair of nodes: $\alpha$, $\rho$, $\phi$, and 
$\delta$. The $\delta$ messages are dependent on the message values from the previous time step and 
the $\phi$ messages are dependent on the next time step, with $\delta_{ij}^t=0$ for all $i,j$ in the 
first time step and $\phi_{ij}^t=0$ for all $i,j$ in the last time step. In a sense, $\delta_{ij}$ and 
$\phi_{ij}$ are symmetrical and for a given factor graph can be interpreted as forward and backward temporal 
smoothing messages. The $\delta_{ij}^t$ messages can be 
written as
\begin{equation}\label{eq_delta}
\delta_{ij}^t= \!\!
\begin{cases}
-\gamma&\text{if $d1\!=\!1,d2\!=\!1$} \\
\rho_{ij}^{t-1}\!+\alpha_{ij}^{t-1}\!-\phi_{ij}^{t-1} \!\!\!\!&\text{if $d1\!=\!1,d2\!=\!0$} \\
-\rho_{ij}^{t-1}-\alpha_{ij}^{t-1}+\phi_{ij}^{t-1} &\text{if $d1\!=\!0,d2\!=\!1$} \\
\gamma&\text{if $d1\!=\!0,d2\!=\!0$},
\end{cases}
\end{equation} 
where $d1=\mathbbm{1}\big( \gamma\geq \rho_{ij}^{t-1}+\alpha_{ij}^{t-1}-\phi_{ij}^{t-1} \big)$ and
$d2 = \mathbbm{1}\big( -\gamma\geq \rho_{ij}^{t-1}+\alpha_{ij}^{t-1}-\phi_{ij}^{t-1} \big)$.

Note that for $\gamma\geq0$, if $d2=1$ then $d1=1$. 
The final $\phi$ messages are similarly derived and become
\begin{equation}\label{eq_phi}
\phi_{ij}^{t-1}= \\
\begin{cases}
-\gamma &\text{if $p1\!=\!1,p2\!=\!1$} \\
\rho_{ij}^{t}+\alpha_{ij}^{t}-\delta_{ij}^{t} &\text{if $p1\!=\!1,p2\!=\!0$} \\
-\rho_{ij}^{t}-\alpha_{ij}^{t}+\delta_{ij}^{t} &\text{if $p1\!=\!0,p2\!=\!1$} \\
\gamma&\text{if $p1\!=\!0,p2\!=\!0$},
\end{cases}
\end{equation} 
where 
$p1 = \mathbbm{1}\big( \gamma\geq \rho_{ij}^{t}+\alpha_{ij}^{t}-\delta_{ij}^{t} \big)$
and $p2 = \mathbbm{1} \big(-\gamma\geq \rho_{ij}^{t}+
\alpha_{ij}^{t}-\delta_{ij}^{t} \big)$. Note that for $\gamma\geq0$, if $p2=1$ then 
$p1=1$. The derivation of $\delta$ and $\phi$ messages using the max sum formulation can be found in the supplementary material.

Finally, let us define the set of exemplars $E$ as
$$E= \{j:\alpha_{jj}^t+\rho_{jj}^t+\delta_{jj}^t+\phi_{jj}^t>0\}.$$
 The exemplar $j'$ for point $i$ is identified as
\begin{equation}\label{eq_exemplar}
j' = \arg\max_{j\in E} \;\alpha_{ij}^t+\rho_{ij}^t+\delta_{ij}^t+\phi_{ij}^t.
\end{equation}

Due to dependence on past and future messages, the EAP message updates are implemented in a forward-backward 
fashion. In each iteration, a message update is performed between the nodes sequentially from the first ($t=1$) to the 
last ($t=T$) time step, followed by a second message update performed backwards from the last time step to the first. 
The number of iterations is determined by a pre-specified maximum number of iterations or algorithm convergence, 
where convergence occurs when exemplar assignment remains static for a number of iterations.

The computational complexity of an EAP iteration (which involves exchanging messages $\alpha,\rho,\delta,\phi$ 
between the nodes in each of $T$ time steps) is $O(N^2T)$, where $N$ is the number of data points. Running an iteration of the classic (static) AP over
$T$ time steps also requires performing $O(N^2T)$ operations. Note that, just as in the case of the classic 
AP, when $N$ is large and the similarity matrix is sparse the messages need not be passed between all pairs 
of points which may significantly reduce complexity.

\subsection{EAP with consensus nodes}
\label{sec:consensusnodes}

While the framework introduced in Section~3.1 enables exchange of messages across different time steps 
and therefore promotes consistency of exemplar selection and cluster structure, tracking clusters presents
challenges; identifying cluster births and death, especially over long time intervals, is particularly difficult. 
For instance, when the exchange of messages via factor nodes $D_{ij}^t$ fails to impose consistency of 
exemplar selection and a data point ends up being an exemplar at time $t$ but not at time $t+1$, it is not 
immediately clear whether the corresponding cluster has died or it is simply represented by a different 
exemplar. In general, answering this question requires computationally costly post-processing. To provide additional stability to the exemplar selection and enable seamless cluster tracking, 
we introduce the concept of consensus nodes -- new variable nodes to be added to the 
graph in Fig.~\ref{figure_EAP} to serve as cluster representatives. Creation of a new consensus node thus
indicates cluster birth, clusters are tracked by observing evolution of the association of data points with 
consensus nodes, and the disappearance of a consensus nodes signals cluster death. Detailed
discussion of these ideas is next. 

{\bf Creating consensus nodes.}
Creation of consensus nodes is done in two stages: first, the forward-backward EAP message exchange 
described in Section~3.1 is conducted until at least two exemplars are identified for each time step; then, 
in the following forward pass, consensus node $i'$ is created for each data point $i$ previously identified 
as exemplar. The feature vector of $i'$ is set to the mean value of the features of all the data points 
having $i$ for exemplar. Messages for 
consensus node $i'$ are initially set equal to those for $i$, with the exception of $\alpha_{i'i}^t$; this 
availability message cannot be initialized by $\alpha_{ii}^t$ since (\ref{eq_alpha}) implies that 
$\alpha_{ii}^t$ may be greater than zero while $\alpha_{i'i}^t$ is at most $0$. To set its initial 
value, we note that $\alpha_{i'i}^t$ can be interpreted as the evidence as to why $i'$ should choose 
$i$ as exemplar and recall the restriction that a consensus node should choose itself as exemplar. Since 
consensus node $i'$ is essentially joining the cluster which contains data point $i$ and is taking over 
the role of exemplar from $i$, it is intuitive that $i$ is the second-best exemplar for $i'$. Let $y$ denote 
the data point that is the second-best exemplar for $i$; we initialize $\alpha_{i'i}^t$ by the evidence as 
to why $i$ should choose $y$ as exemplar. Note that to account for historical exemplar assignments of 
$i$, $y$ should be identified based on message values rather than similarities. Following (\ref{eq_alpha}), 
$\alpha_{ii'}^t$, availability of the new consensus node $i'$ to be exemplar for data point $i$, is initialized 
as $0$ (the maximum value of $\alpha_{jk}^t$ for $j\neq k$). The consensus node created at time $t$ is 
replicated at $t+1$. Assignments of feature vectors and message initialization at $t+1$ follow 
a procedure similar to that at time $t$, with one key difference: since different data points may be the 
most suitable exemplars for a cluster at different time steps, we should not initialize messages for 
$i'$ at $t+1$ using messages for $i$ at $t+1$. Instead, they should be initialized by the messages for 
the most common exemplar at $t+1$ for the data points assigned to exemplar $i$ at time $t$. The 
consensus node creation is formalized by Algorithm \ref{alg:consensus}, where $e_i^t$ denotes the 
exemplar assigned to point $i$ at time $t$ and $x_k^t$ is the feature vector of point $k$ at time $t$. 
Note that since the number of consensus nodes is much smaller than the number of data points $N$, 
the computational complexity of an EAP iteration remains $O(N^2T)$.

\begin{algorithm}[h]
\caption{Cluster birth: Creation of consensus nodes}\label{alg:consensus}
\begin{algorithmic}
\item $V^t \leftarrow$ set of data points at time $t$
\item $C^t \leftarrow$ set of consensus nodes at time $t$
\item $E^t \leftarrow$ set of exemplars at time $t$
\For{$i\in V^t\cap E^t$}:
\State create consensus node $i'$ at time $t$:
\State $x_{i'}^t \leftarrow \frac{1}{\sum_j \mathbbm{1}(e_j^t=i)}\sum_{j:e_j^t=i} x_i^t$
\State initialize message values of $i'$ to those of $i$:
    \For{$j\!\in V^t \cup C^t, m\!\in\! \{\alpha,\rho,\delta,\phi\}$}:
    \State $m_{i'j}^t\leftarrow m_{ij}^t, m_{ji'}^t \leftarrow m_{ji}^t$
    \State $m_{i'i'}^t \leftarrow m_{ii}^t$
    \EndFor
\State update $\alpha_{i'i}^t$ and $\alpha_{ii'}^t$:
\State $y \leftarrow \arg\max_{j\in V^t\setminus i} \alpha_{ij}^t+\rho_{ij}^t+\delta_{ij}^t+\phi_{ij}^t$
\State $\alpha_{i'i}^t \leftarrow \alpha_{iy}^t$, $\alpha_{ii'}^t\leftarrow 0$
\State update exemplars to replace $i$ with $i'$
\EndFor
\item initialize consensus nodes at next time step
\For{$k\in C^t\setminus C^{t+1}$}:
\State $x_k^{t+1} \leftarrow \frac{1}{\sum_j \mathbbm{1}(e_j^t=k)}\sum_{j:e_j^t=k} x_j^{t+1}$ 
\State $l \leftarrow \arg\max_{j\in E^{t+1}}\sum_{i\in V^t} \mathbbm{1}(e_i^t=k)\mathbbm{1}(e_i^{t+1}=j)$
\State set messages for $k$ at $t+1$ using messages for $l$ at $t+1$ following initialization of $i'$ 
messages above
\EndFor
\end{algorithmic}
\end{algorithm}

{\bf Promoting selection of consensus nodes.}
After creating consensus nodes, their selection as exemplars is promoted in order to enable 
efficient cluster tracking. This is accomplished by modifying the definition of factor nodes $D_{ij}^t$ to
\begin{align}
D_{ij}^t(c_{ij}^{t-1},c_{ij}^t) = &\begin{cases}
-\gamma & \!\text{if } c_{ij}^{t-1}\neq c_{ij}^t \\
0 & \!\text{if } c_{ij}^{t-1}= c_{ij}^t=1 \text{ and } j \in C^t \\
-\omega & \!\text{otherwise,}
\end{cases} \label{eq_D}
\end{align}
where $C^t$ denotes the set of consensus nodes and $\gamma\geq\omega\geq0$. Intuitively, modified 
$D_{ij}^t$ encourages temporal smoothness by penalizing changes in cluster memberships and 
rewarding assignments to the nodes in $C^t$. The messages for $\delta_{ij}^t$ are modified accordingly
and become
\begin{equation}\label{eq_delta1}
\delta_{ij}^t= \!\!
\begin{cases}
-\gamma+\omega &\text{if $d1\!=\!1,d2\!=\!1$} \\
\omega\mathbbm{1}(j\in C^t) + \rho_{ij}^{t-1}\!+\alpha_{ij}^{t-1}\!-\phi_{ij}^{t-1} \!\!\!\!&\text{if $d1\!=\!1,d2\!=\!0$} \\
-\rho_{ij}^{t-1}-\alpha_{ij}^{t-1}+\phi_{ij}^{t-1} &\text{if $d1\!=\!0,d2\!=\!1$} \\
\gamma-\omega\mathbbm{1}(j\not\in C^t) &\text{if $d1\!=\!0,d2\!=\!0$},
\end{cases}
\end{equation} 
where $d1=\mathbbm{1}\big( \gamma-\omega\geq \rho_{ij}^{t-1}+\alpha_{ij}^{t-1}-\phi_{ij}^{t-1} \big)$ and
$d2 = \mathbbm{1}\big( -\gamma+\omega\mathbbm{1}(j\not\in C^t)\geq \rho_{ij}^{t-1}+\alpha_{ij}^{t-1}-\phi_{ij}^{t-1} \big)$.
The message updates for $\phi_{ij}^t$ are similarly modified as
\begin{equation}\label{eq_phi1}
\phi_{ij}^{t-1}= \\
\begin{cases}
-\gamma +\omega &\text{if $p1\!=\!1,p2\!=\!1$} \\
\omega\mathbbm{1}(j\in C^t) + \rho_{ij}^{t}+\alpha_{ij}^{t}-\delta_{ij}^{t} &\text{if $p1\!=\!1,p2\!=\!0$} \\
-\rho_{ij}^{t}-\alpha_{ij}^{t}+\delta_{ij}^{t} &\text{if $p1\!=\!0,p2\!=\!1$} \\
\gamma-\omega\mathbbm{1}(j\not\in C^t) &\text{if $p1\!=\!0,p2\!=\!0$},
\end{cases}
\end{equation} 
where 
$p1 = \mathbbm{1}\big( \gamma-\omega\geq \rho_{ij}^{t}+\alpha_{ij}^{t}-\delta_{ij}^{t} \big)$
and $p2 = \mathbbm{1} \big(-\gamma+\omega\mathbbm{1}(j\not\in C^t)\geq \rho_{ij}^{t}+
\alpha_{ij}^{t}-\delta_{ij}^{t} \big)$. Note that for $\gamma\geq\omega\geq0$, if $p2=1$ then 
$p1=1$. As can be seen from (\ref{eq_delta1})-(\ref{eq_phi1}), value of $\delta_{ij}^t$ 
($\phi_{ij}^{t-1}$) depends on the responsibility and availability as well as other temporal smoothing 
messages at the previous (next) time step. Intuitively, if after solving (\ref{eq_exemplar}) $j$ is found to
be an exemplar for $i$, then $\delta_{ij}^t$ ($\phi_{ij}^{t-1}$) attempts to quantify how 
appropriate is the assignment of $j$ as an exemplar in the specified time step while 
accounting for temporal smoothness. The smallest possible value of both 
$\delta_{ij}^t$ and $\phi_{ij}^{t-1} $ is $-\gamma+\omega$, while the largest one is $\gamma$ if 
$j$ is a consensus node and $\gamma-\omega$ if $j$ is a data point; this implies promoting selection of consensus nodes for exemplars, and suggests that by tuning message 
value saturation point one can affect competition between exemplar candidates.

Algorithm \ref{alg:exemplar} formalizes the procedure for identification of exemplars that favors 
consensus nodes. In the algorithm, $e_i$ denotes exemplar for point $i$ and $EC_i$ denotes the set of candidate consensus node exemplars for point $i$.
\begin{algorithm}[h]
\caption{Exemplar identification}\label{alg:exemplar}
\begin{algorithmic}
\item $V^t \leftarrow$ set of data points at time $t$
\item $C^t \leftarrow$ set of consensus nodes at time $t$
\item Identify the set of exemplars $E$:
\item $E\leftarrow \{j:\alpha_{jj}^t+\rho_{jj}^t+\delta_{jj}^t+\phi_{jj}^t>0\}$
\For{$i \in V^t$}
\State $EC_i \leftarrow \{k \in E\cap C^t:\alpha_{ik}^t+\rho_{ik}^t+\delta_{ik}^t+\phi_{ik}^t>0\}$
	\If{$EC_i \neq \emptyset$}
	\State $e_i \leftarrow \arg\max_{k\in EC_i} \alpha_{ik}^t+\rho_{ik}^t+\delta_{ik}^t+\phi_{ik}^t$
	\Else
	\State $e_i \leftarrow \arg\max_{k\in E} \;\alpha_{ik}^t+\rho_{ik}^t+\delta_{ik}^t+\phi_{ik}^t$
	\EndIf
\EndFor \\
\Return $E \leftarrow E\cap \{e_0,\ldots,e_N\}$
\end{algorithmic}
\end{algorithm}
Note that in order to facilitate cluster tracking, an additional update is performed during the exemplar 
identification and assignment. In particular, if consensus node $k$ identifies data point $i$ rather than 
itself as an exemplar, the consensus node takes on messages for $i$ while the data points originally
assigned to $i$ are re-assigned to the consensus node $k$. 


{\bf Disappearance of consensus nodes: cluster death.}
As data evolves and the set of points choosing consensus node $k$ for exemplar varies 
over time, the data vector for $k$ and the similarity matrix are updated. If at some time step
consensus node $k$ is not selected as an exemplar, the cluster 
corresponding to $k$ is considered to have died. Cluster evolution and death is formalized in
Algorithm~\ref{alg:evolution}.
A ``dead" cluster may be ``revived" in future only
in the case of frequent change of exemplars before the message values converge. Suppose 
a consensus node $k$ dies at time $t$ in iteration $n$. If the consensus node $k$ is selected as 
an exemplar at time $t-1$ in iteration $n+1$, it is removed from the set of dead consensus nodes 
at $t$ and its message values at $t$ are re-established according to the initialization in 
Algorithm \ref{alg:consensus}. This process is repeated at $t+1$ if the consensus node $k$ is 
selected as an exemplar at time $t$ in iteration $n+1$. 
\begin{algorithm}[h]
\caption{Cluster evolution and death via consensus nodes}\label{alg:evolution}
\begin{algorithmic}
\item $V^t \leftarrow$ set of data points at time $t$
\item $C^t \leftarrow$ set of consensus nodes at time $t$
\item $\overline{C^t} \leftarrow$ set of dead consensus nodes at time $t$
\item identify exemplars $E^t$
\item \textbf{Cluster evolution}
\For{$k\in C^t\cap E^t$}:
\State $x_k^t \leftarrow \frac{1}{\sum_j \mathbbm{1}(e_j^t=k)}\sum_{j:e_j^t=k} x_j^t$
\EndFor
\item \textbf{Cluster death}
\item $\overline{C^t} \leftarrow \overline{C^t} \cup C^t \setminus E^t$ 
\For{$t'=t+1, \ldots, T$}
\State $\overline{C^{t'}} \leftarrow \overline{C^{t'}} \cup C^t \setminus E^t$ 
\EndFor
\end{algorithmic}
\end{algorithm}

\begin{figure*}[h]
\caption{(a) An illustration of how the consensus nodes are created. (b) An example of EAP that shows 
birth and death of consensus nodes.} 
\label{figure_CNcreation}
\includegraphics[width=4.85in]{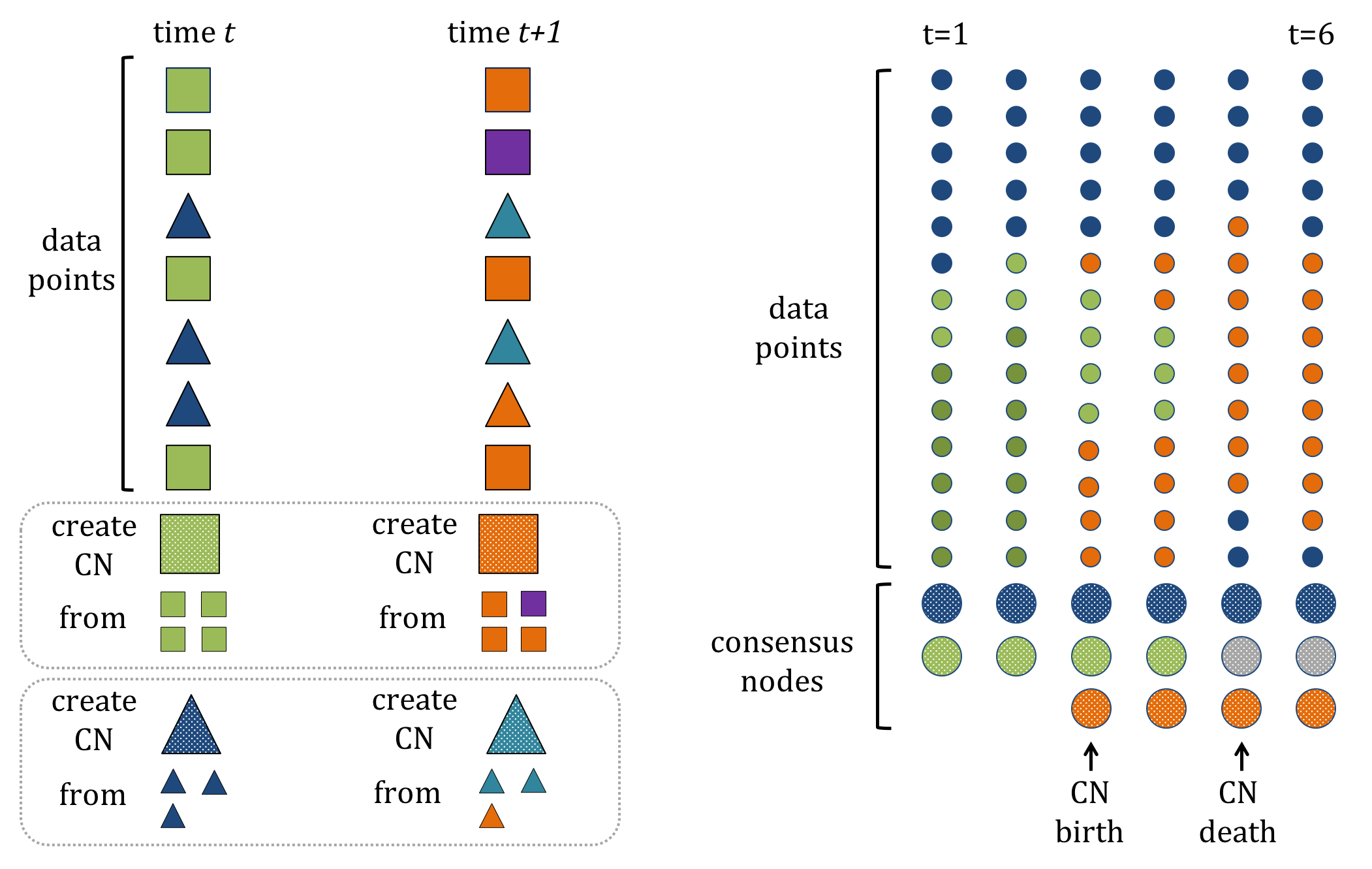}
\centering
\vspace{-0.1in}
\end{figure*}

We conclude the discussion of EAP consensus nodes by an 
illustrative example in Fig.~\ref{figure_CNcreation}. Fig.~\ref{figure_CNcreation}(a) shows the 
process of creating consensus nodes and demonstrates how those nodes enable cluster tracking 
while Fig.~\ref{figure_CNcreation}(b) depicts birth and death of consensus nodes. 
Fig.~\ref{figure_CNcreation}(a) assumes that the basic forward-backward EAP message exchange 
described in Section~3.1 identified two data points as exemplars at time $t$. The cluster 
membership of a data point at time $t$ is indicated by the shape of the polygon depicting the point 
(i.e., the points depicted by triangles belong to one while those depicted by squares belong to the 
other cluster). The color of a polygon represents the corresponding point's exemplar, e.g., ``green" 
is the exemplar for square points while ``blue" is the exemplar for triangular points. The algorithm 
creates two consensus nodes, one for each exemplar. The square consensus node at time $t$ is 
assigned feature vector equal to the mean of the square data points feature vectors and inherits 
message values $\alpha, \rho,\delta,\phi$ from the ``green" exemplar. The triangular consensus 
node at time $t$ has feature vector equal to the mean of the triangular data points feature vectors 
and inherits message values from the ``blue" exemplar. At time $t+1$, three data points were 
identified as exemplars; these exemplars (purple, orange, and teal) may be different from those at 
time $t$ (green and blue) which renders cluster tracking challenging. To provide consistency and
enable cluster tracking, consensus nodes created at time $t$ are replicated at $t+1$ and their 
messages are initialized in a way that acknowledges changes in the features of data points yet 
links clustering solution at $t$ with that at $t+1$. To explain this, note that the shape of the 
polygons representing data points at $t+1$ indicates the structure of clusters at $t$, i.e., the 
squares at $t+1$ indicate data points that were together in one cluster at time $t$ while the triangles 
indicate those that were in the other. To promote consistency of clustering solutions, the square 
consensus node at $t+1$ is assigned feature vector equal to the mean of the square data 
points feature vectors and inherits message values $\alpha, \rho,\delta,\phi$ from the most common 
exemplar for the square data points at $t+1$; in this illustration, the orange exemplar. The triangular 
consensus node at $t+1$ has feature vector equal to the mean of the triangular data points feature 
vectors and inherits message values from the most common exemplar for the triangular data points 
at $t+1$; here, the teal exemplar. After this initialization, the algorithm proceeds to update messages 
including those for the created consensus nodes. Finally, Fig.~\ref{figure_CNcreation}(b) illustrates 
birth of a consensus node (at $t=3$) and death of another (at $t=5$).


{\it Remark:} For particularly noisy data, it might be desirable to restrict creation and preservation 
of consensus nodes to clusters having size above a certain threshold. Setting a threshold may 
aid in finding a stable solution if in the initial iterations many candidate exemplars are identified. Since it is preferred that consensus nodes are chosen as exemplars, limiting the consensus nodes to a 
minimum cluster size discourages assignment of data points to transitive consensus nodes or those that 
are representative of outlier data. Note that such a restriction does not force the final solution to include 
only the clusters larger than the threshold since data points may still emerge as exemplars for small 
clusters, including outliers that result in single-point clusters. 

\subsection{Effect of parameter values} \label{sec:params}

The temporal smoothness and cluster tracking parameters introduced by EAP framework, $\gamma$ and 
$\omega$, affect the number of clusters and the assignment of data points to clusters. In order to 
understand how to set these parameters, it is worth exploring the effects of $\gamma$ and $\omega$ on 
messages $\delta$ and $\phi$ by analyzing definitions \eqref{eq_delta1} and \eqref{eq_phi1}. 

If $\gamma=\omega=0$, it is easy to see that $\delta_{ij}^t=\phi_{ij}^t=0$ for all $i,j,t$ and the solution of 
EAP without consensus nodes would be as same as that obtained by running AP independently at each 
time step.
When $\gamma = \omega > 0$, the $\delta_{ij}^t$ and $\phi_{ij}^t$ messages will be positive if $j$ is a 
consensus node; in fact, these will be the only non-zero messages. Definition of $D_{ij}^t$ in 
\eqref{eq_D} suggests that this setting is equivalent to treating solutions where data points are consistently 
chosen as exemplars in the same way as when a different exemplar is chosen at each time point. When 
$\gamma>0$ and $\omega<\gamma$, the maximum value of $\delta_{ij}^t$ and $\phi_{ij}^t$ is either
$\gamma-\omega$ (if $j$ is a data point) or $\gamma$ (if $j$ is a consensus node). Since consensus 
node $j$ is chosen as exemplar when $\alpha_{ij}^t+\rho_{ij}^t+\delta_{ij}^t+\phi_{ij}^t>0$, $\gamma>0$
makes such a choice more likely.

We conjecture that the value of $\omega$ should be set to a fraction of the value of $\gamma$ (i.e., 
$\omega=\xi\gamma$, $\xi\in[0,1]$); we found that $\xi\in[0,0.5]$ consistently yields good results on real 
datasets. Empirically, high values of $\gamma$ may lead to low number of consensus nodes or clusters. 
This is due to messages $\delta_{ij}^t$ and $\phi_{ij}^t$ taking on a broader range of values which puts 
more emphasis on temporal smoothness in the exemplar detection and assignment. The exemplar selection 
in turn affects creation and survival of consensus nodes. Note that the number of consensus nodes 
created in the first forward pass of Algorithm~1 is typically large and tends to decrease as the message 
values stabilize. This is akin to what is observed in the classic AP algorithm, where fluctuating message 
values lead to many more candidate exemplars in the earlier iterations than in the final solution. We have 
observed that high values of the ratio $\omega/\gamma$ lead to one of two extremes: either too few 
consensus nodes are created in the first pass of Algorithm~1 which then discourages creation of new 
clusters, or too many exemplars are identified which results in too many consensus nodes. Moreover, high 
values of $\omega/\gamma$ make it more likely that the number of consensus nodes does not decrease
with the number of iterations of the algorithm, as we expect, which may result in poor tracking or lead to
higher than optimal number of clusters in the final solution. In the results section, we 
demonstrate the effect of varying $\gamma$ and $\omega$ in experiments on synthetic data sets.

\subsection{Insertion and deletion of time points}\label{sec:indel}

EAP readily handles datasets having points not present for the entire duration of the considered
time horizon; in such scenarios, EAP tracks the set of ``active" data points $V_t$ at each time step 
$t$ and only passes messages between the active data points and active consensus nodes.

Point insertion was previously considered in the context of classic AP with the goal of avoiding to 
perform clustering from scratch every time a new point was introduced in streaming data 
applications \cite{sun_2014}. There, messages for inserted points were initialized using nearest 
neighbors' messages and used in conjunction with the final messages of the AP run on the original 
(smaller) dataset to enable faster convergence of the AP clustering on the new (larger) dataset. 
Although our initialization of consensus node messages is in part motivated by that work, dealing 
with insertions and deletions of points in a temporally evolving dataset requires different treatment. 
In particular, the nearest neighbors of data points present in only a subset of time steps should be 
identified based on both similarities and cluster membership. Moreover, forward-backward nature
of EAP requires initialization of messages in both directions; data points that are deleted (not active 
at $t=T$) can be thought of as insertions in the backwards pass of the algorithm, and should thus 
have their messages updated similar to the insertions in the forward pass. 


In the forward pass of the first iteration, nearest neighbors of inserted points are identified based on 
similarities at the time of insertion. Let $b$ index a data point inserted at time $t$. 
In the first iteration, the 
nearest neighbor of $b$ is $$\text{nn}_b=\arg\max_{j\in B^t}s(b,j)^t,$$ where $s(b,j)^t$ 
is the similarity between $b$ and $j$ at time $t$,
and $B^{t}$ is the set of points active at both $t$ and $t-1$ ($b\not\in B^{t}$). After the first iteration, for data points that 
become active at time $t$, the nearest neighbor is identified as the active data point having the sum of
messages at time $t$ closest in terms of the Euclidean distance. 
More specifically, let $M^t=A^t+P^t+\Delta^t+\Phi^t$, where $A^t,P^t,\Delta^t,\Phi^t$ are matrices consisting of messages $\alpha,\rho,\delta,\phi$, respectively, at time $t$. Let $M_{iB^t}^t$ be the vector 
containing elements from the $i^{th}$ row and $\{k:k\in B^t\}$ columns of $M^t$. Then the nearest 
neighbor of $b$ is
$$\text{nn}_b=\arg\min_{j\in B^{t}} \Vert M_{bB^t}^{t}-M_{jB^t}^{t}\Vert_2.$$ 

Messages of deleted points are initialized in a similar manner during the backwards pass. Let $d$ 
index a point deleted at time $t+1$ and $D^{t}$ be the set of points active at both $t$ and $t+1$ 
($d\not\in D^{t}$). Then the nearest neighbor of $d$ is
$$\text{nn}_d=\arg\min_{j\in D^{t}} \Vert M_{dD^t}^{t}-M_{jD^t}^{t}\Vert_2.$$ 

Once the nearest neighbor is identified, messages of all data points are updated according to the 
EAP update equations. For a data point $b$ inserted at time $t$, we set messages $\delta_{b:}$
and $\delta_{:b}$ to those of the nearest neighbor's, i.e., $\delta_{bj}^t=\delta_{\text{nn}_bj}^t$,
$\delta_{jb}^t=\delta_{j\text{nn}_b}^t$. Similarly, for a data point $d$ deleted at time $t+1$, we set
$\phi_{dj}^t=\phi_{\text{nn}_dj}^t$ and $\phi_{jd}^t=\phi_{j\text{nn}_d}^t$. Note that the other 
messages, $\alpha^t, \rho^t$, and $\phi^t$ for an insertion or $\delta^t$ for a deletion, are updated 
according to the standard EAP message updates rules.


\section{Experimental Results}\label{sec:results}

We test EAP on several synthetic datasets as well as on real ocean, stock and health plan data. 
In our EAP and AP implementations, the similarity between points is defined as the negative squared 
Euclidean distance. For each dataset, the preference (self-similarity) is set to the minimum similarity 
between all data points at a given time. Other AP and EAP hyperparameters are as follows: the maximum number of iterations is 500, and 20 iterations without changes in the exemplars at the last time step are required before declaring convergence. We compare the performance of EAP to that of the 
AFFECT's evolutionary spectral clustering framework \cite{xu_2013} and to the classic (static) AP.
We chose to compare EAP with AFFECT since the latter was shown in \cite{xu_2013} to outperform 
evolutionary k-means \cite{chakrabarti_2006}, evolutionary spectral clustering \cite{chi_2009}, and 
the evolutionary clustering framework in \cite{rosswog_2008} on various synthetic and real datasets. 
The AFFECT framework is implemented using AFFECT Matlab toolbox. For synthetic and ocean 
datasets, AFFECT performs much better if the Gaussian kernel similarity measure is used instead of
the negative squared Euclidean distance (the similarities used by EAP and AP); hence for these
datasets the similarity between points $x_i$ and $x_j$ in AFFECT is defined as 
$\exp(-\|x_i-x_j\|_2^2 / 2\sigma^2)$ where we use the default value $2\sigma^2=5$. The negative 
squared Euclidean distance is used as similarity metric for all algorithms when applied to the 
analysis of the stock dataset. To determine the number of clusters needed to run AFFECT, we used 
the modularity criterion \cite{newman_2006} since it allows for varying number of clusters across time 
and performs better than the alternative approach where the number of clusters is determined by 
maximizing the silhouette width \cite{rousseeuw_1987}. When data labels are available, clustering 
accuracy is evaluated by means of the Rand index \cite{rand_1971} -- the percentage of pairs of 
points correctly classified as being either in the same cluster or in different clusters. Contrary to clustering methods such as k-means which may require many random initializations, there is no random component to AP-based methods and thus all the results in this section are obtained from a single run of the algorithms with the given hyperparameters. The code implementing EAP is at \url{https://github.com/nma14/evolutionary-affinity-propagation}

\subsection{Gaussian data}

We test our algorithm on synthetic datasets generated using four Gaussian mixture models as in \cite{xu_2013}; the Gaussians are $2$-dimensional and datasets consist of $200$ points at each time. The component membership of a point does not change over time unless specified otherwise. For the first dataset, the Gaussian distributions are well-separated. In the colliding Gaussians dataset, means of the Gaussians are nearing in the first 9 time steps, and remain static for the remaining time. The last two datasets consist of points that change cluster membership and are generated similarly to the colliding Gaussians setting. The difference is that in the third dataset, some points change clusters at $t=10,11$, while in the fourth dataset, some points switch to a new, third, Gaussian component at $t=10,11$. More details on the synthetic datasets can be found in the supplementary material.

In addition to the AFFECT framework, EAP is compared to clustering with AP independently at each 
time step. Moreover, to demonstrate the impact of consensus nodes, we also test an EAP 
implementation that does not employ consensus nodes and has message updates equivalent to 
those of EAP with $\omega=0$ (basically, the EAP framework in Section~3.1). The damping factor ($\lambda$) for AP and the EAP implementations is 0.9. The results of EAP 
($\gamma=2$ and $\omega=1$) and its no-consensus-nodes variant labeled EAP:noCN ($\gamma=2$) are shown 
in Table \ref{table_rand}. The corresponding numbers of distinct exemplars across the time steps are 
shown in Table \ref{table_exemp}. The number of exemplars for AFFECT is not included in Table \ref{table_exemp} since AFFECT is not an exemplar-based clustering algorithm and requires a predefined number of clusters as an input to the algorithm.

\begin{table}
\centering
\renewcommand{\arraystretch}{1}
\caption{Accuracy of clustering Gaussian datasets in terms of Rand index.}
\label{table_rand}
\begin{tabular}{cccccc}
\hline
Dataset & EAP & AP & EAP:noCN & AFFECT \\
\hline
separated Gaussians & 1 & 0.768 & 0.934 & 1\\
colliding Gaussians & 1 & 0.943 & 0.986 & 1\\
cluster change & 0.997 & 0.879 & 0.991 & 0.955\\
third cluster & 0.995 & 0.971 & 0.992 & 0.963\\
\hline
\end{tabular}
\end{table}
\begin{table}
\centering
\renewcommand{\arraystretch}{1}
\caption{Inferred numbers of distinct exemplars (mean number of clusters) when clustering Gaussian datasets.}
\label{table_exemp}
\begin{tabular}{cccc}
\hline
Dataset & EAP & AP & EAP:noCN \\
\hline
separated Gaussians & 2 (2) & 111 (3.93) & 17 (2.68) \\
colliding Gaussians & 2 (2) & 48 (2.20) & 11 (2.12) \\
cluster change & 2 (2) & 89 (2.40) & 12 (2.04)\\
third cluster & 3 (2.64) & 101 (2.68) & 17 (2.68)\\
\hline
\end{tabular}
\end{table}

EAP achieved near-perfect clustering and correctly tracked clusters for all 4 datasets. Clustering 
with EAP messages but without consensus nodes yielded more accurate results and tracked 
clusters better than when clustering was done with AP independently at each time step. However, 
accuracy of EAP:noCN was lower than that of EAP on most datasets due to the former's struggle 
when dealing with competing exemplars at consecutive time steps. Specifically, when several 
data points are good exemplar candidates for a cluster, message dependence on the future and 
past time steps may result in more candidate exemplars at a single time step than if the clustering 
were performed solely based on the data collected in that time step. Thus despite the Rand index 
of EAP:noCN being higher than that of AP, competing exemplars that the former needs to deal with 
may lead to a higher than actual number of clusters. The inclusion of consensus nodes and 
message initializations and updates in EAP overcome this limitation of EAP:noCN. In the dataset 
where points form a new third cluster, EAP:noCN slightly outperforms EAP in terms of clustering accuracy 
at individual time steps (Table \ref{table_rand}); however, EAP:noCN performs poorly at tracking 
clusters across time since it finds $17$ distinct exemplars compared to EAP which correctly 
identifies $3$ clusters and tracks them perfectly (Table \ref{table_exemp}). EAP outperforms 
AFFECT with spectral clustering on the datasets having points changing clusters and points 
forming the third cluster, providing significantly higher accuracy following cluster 
membership change. Note that AFFECT does not detect the third cluster, formed at $t=10$, until 
$t =18$; as a consequence, its Rand index is lower than that achieved by AP applied to data at each 
time instant independently. 
\begin{figure}[h] 
\centering
\vspace{-0.1in}
\caption{Rand index for EAP, EAP:noCN, AP and AFFECT applied to clustering colliding Gaussians 
with the change in cluster membership (left) and the appearance of a third cluster (right).} 
\label{figure_gauss}
\includegraphics[width=3in]{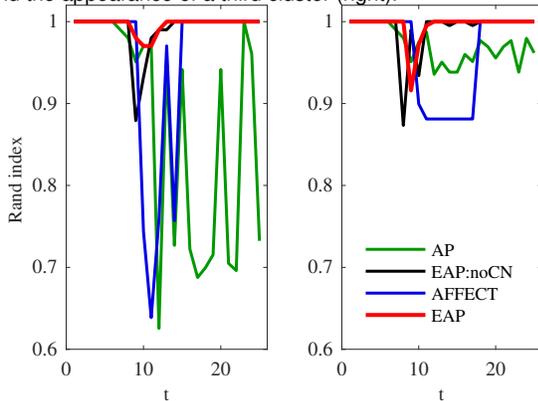}
\centering
\vspace{-0.1in}
\end{figure}
Figure \ref{figure_gauss} shows the Rand index for the colliding Gaussians with some points changing 
cluster membership at times $t\in\{10,11\}$ (left panel) and colliding Gaussians with some points switching 
to a new third cluster at times $t\in\{10,11\}$ (right panel). Note that as the data points from the two 
Gaussians get closer, the Rand index for AP (green) starts deteriorating. When the third cluster is introduced 
or when data points switch cluster membership, the Rand index for AP drops significantly. AP's low
Rand index in those settings is due to close proximity of some
data points from one cluster to the points in another cluster; AP cannot rely on data points history to correct 
the clustering error. Note that in evolutionary clustering without consensus nodes (EAP:noCN, black)
some exemplars may compete against each other and the number of clusters may be overestimated, 
resulting in a lower Rand index for $t=9$ and the following time steps. EAP (red) yields the best clustering results on the
dataset with points changing cluster membership; it exhibits a slight decrease in the Rand index around 
the time of perturbation in cluster memberships, followed by a quick recovery. In application
to the dataset where points suddenly form a third cluster (see Fig.~\ref{figure_gauss}), EAP yields perfect clustering 
at all times except for a brief period ($t\in\{9, 10,11\}$) that starts when the new cluster is formed.

\vspace{-0.1in}
\begin{figure}[h] 
\centering
\caption{Rand index for EAP applied to clustering colliding Gaussians 
with the change in cluster membership (left) and the appearance of a third cluster (right) for combinations of $\gamma$ and $\omega$.} \label{figure_hyperparams}
\includegraphics[width=3.5in]{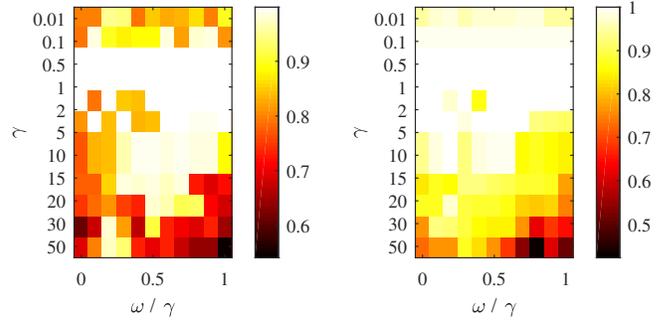}
\centering
\vspace{-0.15in}
\end{figure}

The values chosen for $\gamma$ and $\omega$ were empirically shown to be robust in applications of EAP 
on various data sets. Here we examine in more details how tuning these parameters affects clustering 
results. In Fig.~\ref{figure_hyperparams} we show the performance of EAP on two of the synthetic 
datasets: colliding Gaussians with a fraction of data points changing clusters (left) and the appearance of a third 
cluster (right) 
over a range of $\gamma$ and $\omega/\gamma$ values. In general, a large difference between $\gamma$ 
and $\omega$ (lower left corner of the plots) results in a higher number of clusters in the solution since 
temporal smoothness is valued more than having consensus node exemplars; high values of $\gamma$ and $\omega$ 
(lower right corner of plots) result in a lower number of clusters since data points are discouraged from 
both switching clusters and choosing new data point exemplars. The plot on the right implies that creation 
of a new cluster may be readily identified using a wide range of parameter values, whereas 
increasing $\gamma$, which encourages temporal smoothness, may render the problem in settings where 
data points switch membership between colliding clusters challenging. Through the selection of $\gamma$ 
and $\omega$, the user has the ability to gear the clustering solution towards temporal stability or the 
discovery of new clusters.

\subsection{Real (experimental) data} 

We tested the proposed evolutionary clustering algorithm on three real data sets: ocean temperature 
and salinity data at the location where the Atlantic Ocean meets the Indian Ocean, stock prices 
from the first half of year 2000, and medication adherence star ratings for Medicare health plans.

\subsubsection{Ocean water masses}

Argo, an ocean observation system, has been tracking ocean temperature and salinity since 2000. 
More than 3900 floats currently in the Argo network cycle between the ocean surface and 2000m 
depth every 10 days, collecting salinity and temperature measurements. The primary goal of the Argo 
program is to aid in understanding of climate variability. Evolutionary clustering provides a way to 
discover and track changes in water masses at different depths. A water mass is a body of water 
with a common formation and homogenous values of various features, such as temperature and 
salinity. Study of water masses can provide insight into climate change, seasonal 
climatological variations, ocean biogeochemistry, and ocean circulation and its effect on transport 
of oxygen and organisms, which in turn affects the biological diversity of an area. Clustering has 
previously been used to identify water masses with datasets comprising temperature, salinity, 
and optical measurements \cite{oliver_2004, 
chen_2011, qi_2014}. However, in studies that explored seasonal variations of water masses, 
data at different time points of interest are analyzed independently and then compared in order 
to find variations \cite{chen_2011,qi_2014}. 
\begin{figure}
\vspace{-0.05in}
\caption{Water masses at 1000 dbar. Clustering by EAP (first row), EAP without consensus 
nodes (second row), AP (third row), and AFFECT (bottom row) at $t=2$ and $t=3$. Cluster
membership of each data point is color-coded, visualizing evolution of clusters across time.} 
\label{fig_ocean}
\includegraphics[width=3.5in]{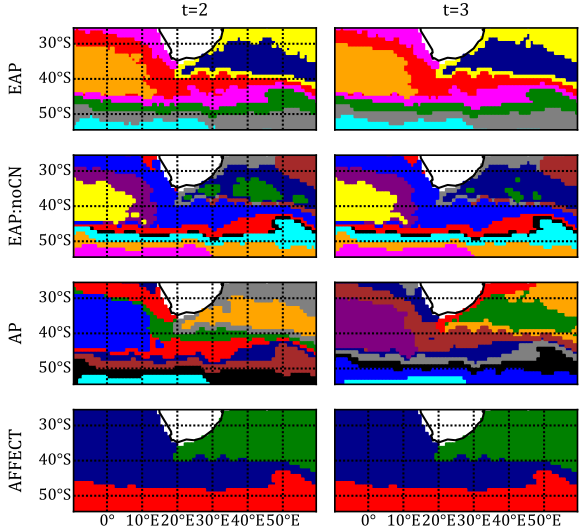}
\centering
\vspace{-0.05in}
\end{figure}

We examine data from the Roemmich-Gilson (RG) Argo Climatology \cite{roemmich_2009} 
which contains monthly averages (since January 2004) of ocean temperature and salinity data 
with a 1 degree resolution worldwide. Clustering is performed on the temperature and salinity 
data at the location near the coast of South Africa where the Indian Ocean meets the South 
Atlantic. Specifically, the data is obtained from the latitudes 25$^{\circ}$ S to 55$^{\circ}$ S and 
longitudes 10$^{\circ}$ W to 60$^{\circ}$ E. The feature vectors used to determine pairwise 
similarities contain the monthly salinity and temperature from April to September, the Austral 
winter, acquired starting in the year 2005 ($t=1$) until 2014 ($t=10$). Temperature and salinity 
were normalized by subtracting the mean and dividing by the standard deviation of the entire 
time frame of interest. EAP is performed with $\gamma=2$, $\omega=1$, $\lambda=0.9$. 

The results for EAP, EAP without consensus nodes (EAP:noCN), AP, and AFFECT with spectral 
clustering are shown for years 2006 and 2007 ($t=2,3$) in Figure \ref{fig_ocean}. EAP tracks 
clusters of water masses across time, where the colors in the top 2 panels of Figure \ref{fig_ocean} 
are indicative of distinct exemplars. For instance, EAP assigned the same exemplar to the green 
area at $t=2$ and the green area at $t=3$. Several distinct water masses are clearly identifiable. 
EAP:noCN is similarly able to track some of the clusters, but some water masses are further 
divided. The colors in the AP plots indicate different clusters per time step, but are not related
across time since the AP results for $t=2$ and $t=3$ do not have any exemplars in common. AP 
identifies many more clusters than EAP, most of which cannot be unambiguously related across 
time. AFFECT identifies only 3 clusters, grouping together known water masses that EAP is able 
to distinguish.

The temperature and salinity averages for the water masses clustered by EAP at $t=2$ 
are shown in Table \ref{table_ocean}. A combination of these values and the geographic 
location of the clusters suggests several known water masses are identified by EAP. In
particular, the yellow cluster in the top panels of Figure 
\ref{fig_ocean} represents the water around the Agulhas currents \cite{talley_indian, 
rusciano_2012}, which reaches down to the ocean bottom and is one of the strongest 
currents in the world. The cyan and gray clusters likely correspond to the Lower Circumpolar Deep 
Water (LCDW) and the Upper Circumpolar Deep Water \cite{talley_southern}. 
These two clusters are characterized by higher salinity and lower temperature than the 
other clusters, in agreement with literature about the water masses \cite{talley_southern} where the LCDW characterized as a salinity maximum layer with previously measured salinity near 34.6-34.74 and temperature near 0.5-1.8$^{\circ}$ C in the mapped region. The green, red, and orange clusters are likely components of the Antarctic 
Intermediate Water (AAIW), which is characterized by lower salinity than other water masses 
\cite{talley_atlantic}. Salinity of the AAIW at 1000m 10$^{\circ}$ C west of the mapped area has been previously found at 34.3-34.5 \cite{talley_atlantic}. Though the clusters may all belong to the AAIW, such a large water 
mass may be divided into regional varieties, where, for example, the green cluster may be 
indicative of the Atlantic AAIW near the Subantarctic Front \cite{rusciano_2012}.

\begin{table}
\centering
\renewcommand{\arraystretch}{1}
\caption{Salinity and temperature of the water masses identified by EAP.}
\label{table_ocean}
\begin{tabular}{ccc}
\hline
Cluster color & salinity & temperature ($^{\circ}$ C) \\
\hline
yellow &  34.49&  6.02\\
blue &  34.53& 7.03 \\
red &  34.40& 3.90\\
orange &34.31 & 3.47\\
magenta & 34.46 & 3.13 \\
green & 34.58& 2.53\\
gray & 34.70&2.14\\
cyan & 34.71& 1.17\\
\hline
\end{tabular}
\end{table}

\subsubsection{Stock prices}\label{sec:stockresults}

Evolutionary clustering can provide insight into the dynamics of stocks and can be an alternative to examining 
a backward-looking covariance matrix using monthly stock returns to identify stocks that behave similarly or 
differently. For example, by choosing different time lengths and resolutions when constructing feature vectors, 
evolutionary clustering can provide insight into groups of stocks behaving similarly during a market regime 
switch, such as a bubble bursting. These methods can also be used in portfolio diversification, to ensure the 
stocks in a portfolio do not cluster together despite diversification by sector or industry.

Daily closing stock prices from January to June 2000 for 3424 stocks were obtained from the CRSP/Compustat 
Merged Database \cite{crsp}. The stocks can be divided into 10 groups based on the S\&P 500 Global Industry 
Classification Standard (GICS) sectors. Such sectors comprise energy, health care, financials, information 
technology, and utilities. The time period was chosen to include the dot-com bubble burst in March of 2000 -- stock prices peaked on March 10, followed by a steep decline. Table \ref{table_sectors} shows the sectors and number of stocks included in the analysis. More details about this data can be found in the supplementary material.

\begin{table}
\centering
\renewcommand{\arraystretch}{1}
\caption{The number of stocks by sector.}
\label{table_sectors}
\vspace{-0.05in}
\begin{tabular}{ccc}
\hline
&Sector & no. of stocks \\
\hline
E&energy  & 204 \\
M&materials& 241 \\
I&industrials & 523 \\
D&consumer discretionary & 576 \\
S&consumer staples & 171 \\
H&health care& 376 \\
F&financials & 642 \\
IT&information technology & 499 \\
T&telecommunications & 61 \\
U&utilities & 131\\
\hline
\end{tabular}
\end{table}

We clustered stocks using EAP, AP, the AFFECT framework with spectral clustering, and static spectral 
clustering. Unlike the clustering results on synthetic data in Section~4.1, the methods yielded different number
of clusters. Since the Rand index is biased towards solutions with a high number of clusters, we also provide 
the results for the modified Rand index \cite{givoni_2009ss, arzeno_2015} defined as
\begin{align}
&\text{modRand} =  \\ &\frac{\sum\limits_{i>j} \mathbbm{1}(c_i=c_j)\mathbbm{1}(\hat{c}_i=\hat{c}_j)}{2\sum\limits_{i>j} \mathbbm{1}(\hat{c}_i=\hat{c}_j)} \nonumber + \frac{\sum\limits_{i>j} \mathbbm{1}(c_i\neq c_j)\mathbbm{1}(\hat{c}_i\neq \hat{c}_j)}{2\sum\limits_{i>j} \mathbbm{1}(\hat{c}_i\neq \hat{c}_j)}.
\end{align}
In the modified Rand index, with values ranging from 0 to 1, the pairs of points that are correctly identified as 
being in different clusters can account for only half of the score, diminishing the bias towards solutions with a 
large number of clusters. The average clustering results for the 6 months using the sectors as the ``true" 
labels are presented in Table \ref{table_stockresults}, where the number of clusters for AFFECT was chosen 
using the modularity criterion and the number of clusters for static spectral clustering was set to 10. For algorithms without a predefined number of clusters, the number of clusters is presented as a range, corresponding to the minimum and maximum number of clusters found across time steps. EAP was
run with $\gamma=5$, $\omega=1$, $\lambda=0.9$, and the threshold on the minimum consensus node cluster size was 
set to 20. In the results, all clusters with 20 or more data points were associated with consensus node exemplars 
while smaller clusters that were part of the solution were associated with data point exemplars. 

\begin{table}
\centering
\renewcommand{\arraystretch}{1}
\caption{Results of Clustering Stock Data}
\label{table_stockresults}
\vspace{-0.05in}
\begin{tabular}{cccc}
\hline
Algorithm & Rand & modRand & no. of clusters \\
\hline
EAP & 0.858 & 0.562 & 50-67 \\
AFFECT & 0.799 & 0.515 & 10 \\
AP & 0.861 & 0.530 & 107-115 \\
spectral & 0.797 & 0.509 & 10 \\
\hline
\end{tabular}
\end{table}

EAP achieves a higher modified Rand index than AFFECT, static AP, and static spectral clustering. 
The EAP solution has between 50 and 67 clusters at each time step, 18 of which are common to all time steps. 
This indicates that most clusters are active for only a subset of time, which would make the 
cluster-matching task performed automatically by EAP a challenging post-processing step for other 
static or evolutionary clustering algorithms. Note that the number of clusters should not be expected 
to match the number of sectors. Such a solution, as it happens to arise when using AFFECT, yields 
highly mixed clusters containing significant fractions of stocks from variety of sectors. We refer to 
a cluster as being dominated by a certain sector if the sector contributes at least twice as many 
stocks to the cluster as any other sector. Results of AFFECT suggest that the financials 
sector dominates one cluster each month, the information technology sector dominates one cluster 
from April to June, and the energy sector dominates one cluster in February and March. Assuming 
the clusters dominated by the information technology sector are in fact snapshots of a single cluster
evolving over time, examination of the information technology stocks in those clusters shows that 15 
stocks are present in the cluster from April to May and 3 stocks are present in both May and June. The highly fluctuating cluster membership may indicate that the clustering solution should 
have had a higher number of clusters, where some clusters may be stable across time (corresponding 
to either particular sectors or general market trends) while others may experience fluctuating 
memberships. Interestingly, the AFFECT framework with spectral clustering that uses the number of 
clusters as identified by EAP yields a modified Rand index of $0.544$ (i.e., higher than $0.515$
achieved by setting the number of clusters according to the AFFECT's modularity criterion). Evidently, 
EAP's strategy for determining the number of clusters leads to more accurate clustering 
solutions, and ultimately allows EAP to precisely detect cluster birth and track their evolution.

The EAP solution contains clusters dominated by the information technology, financials, energy, utilities, 
materials, consumer discretionary, and consumer staples sectors, with the cluster-tracking ability of EAP 
signaling some of these clusters are dominated by a given sector at multiple time points. The two clusters 
dominated by the energy sector and the cluster dominated by the utilities sector would likely not be 
identified in a solution with a low number of clusters since these industries combined correspond to 
less than 10 percent of the stocks in the analysis. Additionally, 5 of the 18 clusters that are active at 
all time steps are dominated by a sector at all time steps (2 information technology, 1 financials, 1 energy, 
1 utilities). As in the case of AFFECT, cluster memberships change over time, with between 42 and 
76 percent of the stocks remaining in the same cluster through consecutive months. 

\begin{figure}
\caption{EAP cluster membership by sector. Sectors are on the horizontal axis, clusters are along the vertical axis. A color indicates participation of a sector in a given cluster in terms of 
percentage of the cluster's size, with red corresponding to higher percentage. Cluster births and deaths 
lead to emergence and deletion of rows, respectively; blank rows indicate clusters that are not active. 
Sector labels are defined in Table \ref{table_sectors}.} \label{figure_stockmemb}
\vspace{0.05in}
\includegraphics[width=3.5in]{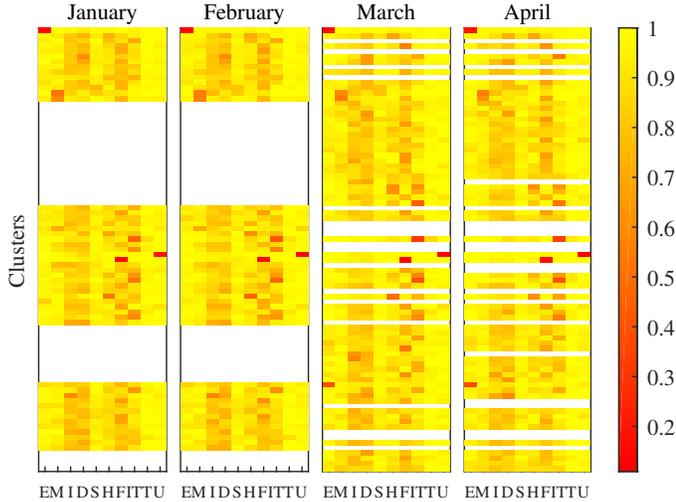}
\centering
\end{figure}

A significant change in the structure of clusters occurs between February and March, the months containing 
55 and 67 clusters, respectively. In Fig. \ref{figure_stockmemb}, each panel corresponds to a time 
step in EAP, the horizontal axis corresponds to the sectors (Table \ref{table_sectors}), and each row 
corresponds to a specific cluster tracked across time, with blank rows indicating inactive 
clusters. The color represents the percentage of the cluster that belongs to a given sector, with red 
indicating higher percentage. The active clusters undergo a major change between February and March 
and remain more consistent between March and April, suggesting a reorganization in March. 
The sector-dominated clusters can be identified in Fig. \ref{figure_stockmemb} by observing locations 
of red rectangles, and their dynamics can be tracked. For instance, it can be seen that the clusters 
dominated by utilities (U) and energy (E) sectors are consistent across time, and that the cluster 
re-organization in March results in a second energy-dominated cluster whose structure remains preserved
through April. The numerous cluster births and deaths illustrated in Fig. \ref{figure_stockmemb} further 
emphasize advantages of automatic cluster number detection and cluster tracking that are among EAP's
features.

\subsubsection{Medicare star ratings}
The Centers for Medicare and Medicaid Services (CMS) evaluates Medicare plans via a 5-star system, where high star ratings result in bonuses while low star ratings may lead to plan termination. The continually rising cost of healthcare and the rating system that provides financial incentives motivate efforts to identify classes of health plans and study their evolution. To form an overall star rating for a Medicare plan, CMS evaluates various aspects of the plan including measures encompassing health screenings and tests, management of chronic conditions, hospital readmission rate, customer service, and member experience. The overall star rating for the plan is a weighted average of the star ratings for the individual measures. The ratings both provide accountability for the health plans and help consumers select a plan. 

We consider diabetes, hypertension, and cholesterol medication adherence data from Medicare Advantage plans with a prescription drug plan in 2012-2016 \cite{cmsdata}. The data is available as both the raw score (0-100) and the star rating assigned to each individual measure. The medication adherence raw score indicates the percentage of adherent members in the measurement period, e.g., a year. 
To be included in the study, the raw data had to be available for at least 2 years between 2012 and 2016, with at least one score available for each of the 3 categories of medication adherence. The medication adherence data is reported on a yearly basis ($T=5$). The final dataset consisted of 597 MA plans from 45 states,  the District of Columbia, and Puerto Rico. Three-dimensional feature vectors collecting medication adherence raw scores were constructed for each plan and each time point.

We cluster the data using EAP implemented with $\omega=1,\gamma=2$, $\lambda=0.5$, and a minimum cluster size of $20$ for consensus node creation. The data were also clustered with AP and EAP without consensus nodes (EAP:noCN), using the same parameters as EAP where applicable, and with AFFECT employing the modularity criterion to determine the number of clusters.

\begin{figure}[!h]
\begin{centering}
\includegraphics[width=3.5in]{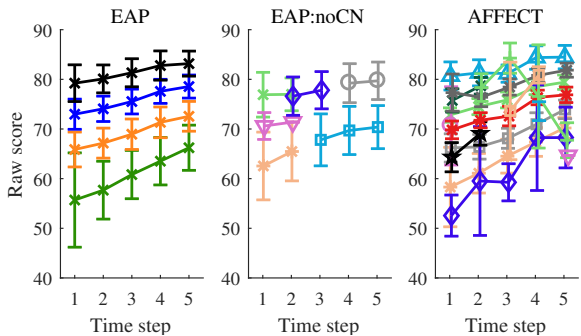}
\caption{Average non-imputed raw medication adherence scores per cluster for years 2012-2016 for EAP (left), EAP without consensus nodes (middle), and AFFECT (right). }
\label{fig_stars_base}
\end{centering}
\end{figure}

EAP revealed 4 clusters that track throughout all 5 time steps. We refer to the clusters by numbers (clusters 1-4), indicating ranking of the average raw scores of the cluster members from worst to best (Fig. \ref{fig_stars_base}).
Most plans ($90-95\%$) remain in the same cluster across consecutive time steps, and plans that do change cluster membership typically switch between adjacent clusters (e.g., from cluster 3 to 2 or 4). 
All clusters in EAP have an upward trend in medication adherence scores, suggesting that inclusion of such a measure in the CMS star ratings is having the intended effect. The average adherence score of cluster 2 in 2016 ($t=5$) is similar to that of cluster 3 in 2012 ($t=1$); a similar relationship exists between clusters 3 and 4. We note that cluster 1 has the greatest improvement
in medication adherence raw scores while cluster 4 has the
least improvement over time. The large improvement in
cluster 1 is likely due to a combination of the survival of
the fittest effect, where some the poorly performing plans
disappear, and a greater overall improvement in adherence
measures than other clusters. The smaller improvement in
cluster 4 suggests there is a ceiling on a realistic maximal
value of medication adherence.
Inspection of the plans in each cluster reveals plan
type, geographical, and parent company differences between
clusters. Further discussion can be found in the supplementary material.

\begin{table}
\renewcommand{\arraystretch}{1}
\caption{Number of Clusters per Year}
\label{table_clustNum}
\vspace{-0.1in}
\centering
\begin{center}
\begin{tabular}{cccccc}
\hline
 & 2012 & 2013 & 2014 & 2015 & 2016 \\
\hline
EAP &  4 & 4 & 4 & 4 & 4\\
\hline
EAP:noCN & 3 & 4 & 2 & 2 & 2\\
\hline
AP & 4 & 98 & 203 & 3 & 292\\
\hline
AFFECT & 10 & 10 & 9 & 10 & 10\\
\hline
\end{tabular}
\end{center}
\end{table}

Clustering with EAP:noCN, AP, and AFFECT is characterized by frequent cluster membership changes; the latter two typically lead to unreasonably high number of clusters. Table \ref{table_clustNum} displays the number of clusters found at each time step for each algorithm. The low number of clusters found by EAP:noCN does not reflect stability of the solution since up to 72\% of the plans changed cluster membership at consecutive time steps. With AFFECT, up to 51\% of plans change cluster membership at consecutive time steps. Both of these produce clusters that exist for only one or two time steps (Fig. \ref{fig_stars_base})). Additionally, clusters formed by AFFECT are largely overlapping (Fig. \ref{fig_stars_base}) even when plotted by the individual medication components (not shown). These results indicate that EAP is capable of identifying stable clusters and tracking them over time while other evolutionary clustering algorithms provide less informative solutions.

\begin{section}{Future Directions}\label{sec:future}

A key limitation of EAP, partially inherited from traditional affinity propagation, is a restriction on the shape of 
clusters that are inferred by the algorithm. We do not expect that EAP will be able to accurately identify spiral 
clusters or clusters on a manifold, although we did not test such scenarios. On a different note, assignment of a 
consensus node's data values as the mean of the data values of the cluster members performed well with the 
use of negative Euclidean distance as the similarity. However, traditional AP has shown the ability to also work 
well with non-metric similarities such as correlation. As part of the future work, it would be beneficial to 
determine how the data should be updated at consensus nodes for an assortment of similarities.

An interesting potential extension of the EAP methodology involves reinterpretation of time 
points. Data is not required to be of the same type at different time points as long as it can be tracked 
and the similarity between data points can be calculated at each time step. Instead of clustering data at different 
time points, a similar methodology could be applied when considering characterization of an instance by 
different types of data. For example, in a medical settings, patients can be clustered given different sections of 
the medical record such as vital signs, lab test results, and some qualitative measures, where EAP 
would encourage patients to belong to the same cluster in different modalities while still allowing for membership in 
different segments.

\end{section}

\begin{section}{Conclusion}\label{sec:conclusions}

We developed an evolutionary clustering algorithm, evolutionary affinity propagation (EAP), 
which groups points by passing messages on a factor graph. The EAP graph includes factors 
connecting variable nodes across time, inducing temporal smoothness. We introduce the 
concept of consensus nodes and describe message initialization and updates that encourage 
data points to choose an existing consensus node as their exemplar. Through these nodes, 
we can identify cluster births and deaths as well as track clusters across time. EAP 
outperforms an evolutionary spectral clustering algorithm as well as the individual time step 
clustering by AP on several Gaussian mixture models emulating circumstances such as 
changes in cluster membership and the emergence of an additional cluster. When applied to 
an ocean water dataset, EAP was able to identify known water masses and automatically match 
the discovered clusters across time. In stock clustering and health plan clustering applications, EAP yields a more 
accurate and interpretable solution than existing static and evolutionary clustering methods.
EAP's capability to identify the number of clusters and perform cluster tracking without additional pre- or post-processing steps makes it a desirable algorithm for studying the evolution of clusters 
when data is acquired at multiple time steps, such as in the study of climate change or 
exploration of social networks.

\end{section}

\ifCLASSOPTIONcompsoc
  \section*{Acknowledgments}
\else
  \section*{Acknowledgment}
\fi

This work was supported by the NSF Graduate Research Fellowship grant DGE-1110007. The 
authors would like to thank Isabella Arzeno for her oceanography advice on water mass 
clustering, ranging from the suggestion of the problem and dataset to the interpretation of the 
results.

\ifCLASSOPTIONcaptionsoff
  \newpage
\fi

\bibliographystyle{IEEEtran.bst}


%

\begin{IEEEbiography}[{\includegraphics[width=1in,height=1.25in,clip,keepaspectratio]{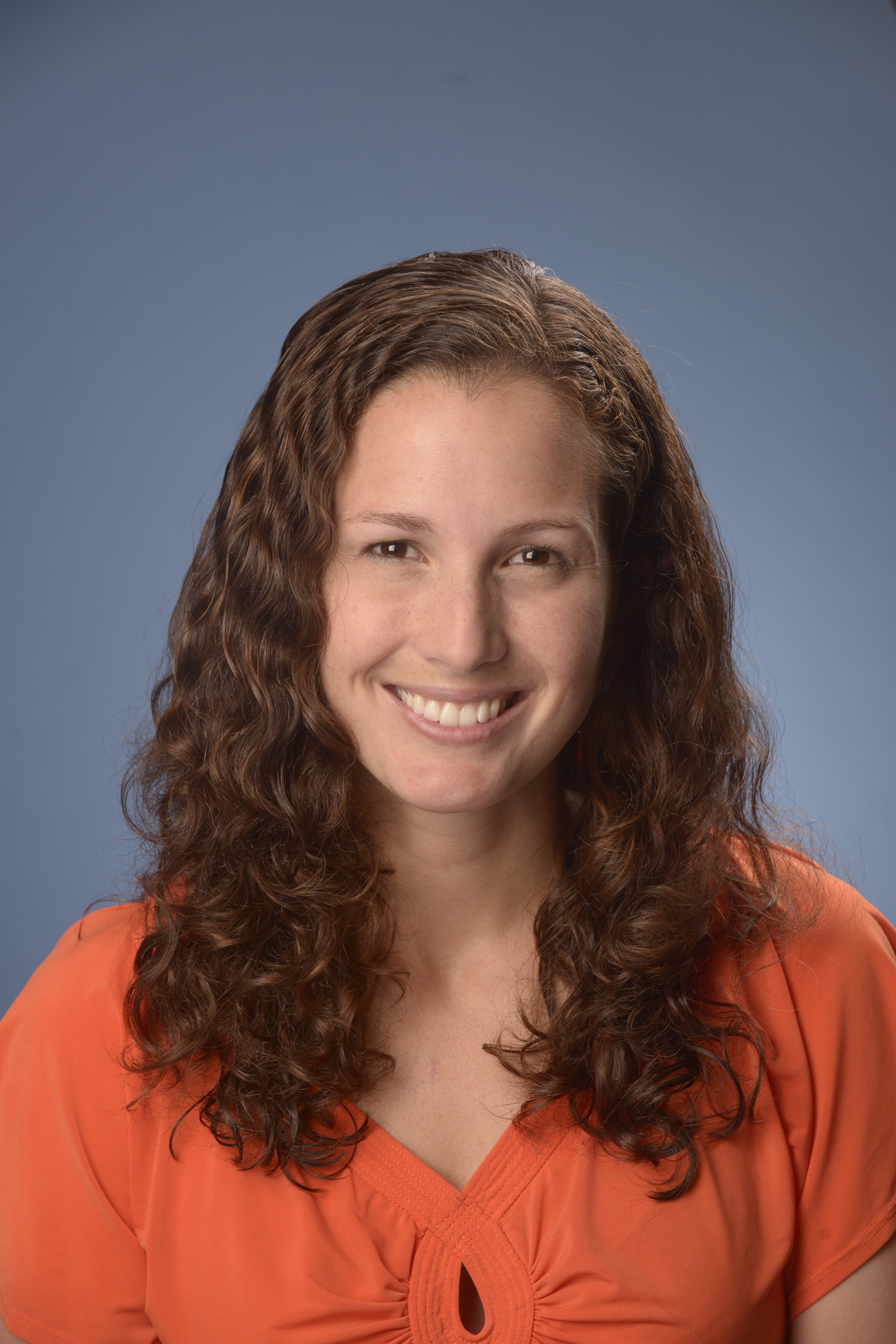}}]
{Natalia M. Arzeno} received the B.S. (2006) and M.Eng. (2007) degrees from the
Massachusetts Institute of Technology, and the Ph.D. degree (2016) from the University
of Texas at Austin, all in electrical engineering. She is a National Science 
Foundation Graduate Research Fellowship and Donald D. Harrington Dissertation Fellowship recipient. Her research interests include machine learning, 
data mining, and healthcare analytics.
\end{IEEEbiography}

\begin{IEEEbiography}[{\includegraphics[width=1in,height=1.25in,clip,keepaspectratio]{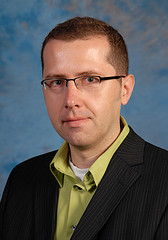}}]{Haris Vikalo} received the Ph.D. in electrical engineering from Stanford University in 2003. From January
2003 to July 2003 he was a Postdoctoral Researcher, and from July
2003 to August 2007 he was an Associate Scientist at the California
Institute of Technology. Since September 2007, he has been with the
Department of Electrical and Computer Engineering, the University of
Texas at Austin. His research interests include signal processing, machine learning, bioinformatics, and communications.
\end{IEEEbiography}




\pagebreak

\twocolumn[{\centering{\Huge  Supplementary Material for ``Evolutionary Clustering via Message Passing"}
\vspace{0.2in}}]

\setcounter{equation}{0}
\setcounter{figure}{0}
\setcounter{table}{0}
\setcounter{page}{1}
\setcounter{section}{0}
\renewcommand{\thesection}{S\arabic{section}}
\renewcommand{\theequation}{S\arabic{equation}}
\renewcommand{\thefigure}{S\arabic{figure}}
\renewcommand{\thepage}{S\arabic{page}}
\renewcommand{\thetable}{S\arabic{table}}

In this supplement to ``Evolutionary Clustering via Message Passing," we present detailed derivations of the $\delta$ and $\phi$ messages in EAP, specifics on the data vectors creation for some of the experiments, and a more thorough description of data and analysis of results obtained by applying EAP to clustering of the Medicare star ratings dataset.

\section{Message derivations: $\delta$ and $\phi$}
In this section, all equation references correspond to the main text.

The $\delta_{ij}^t$ are derived as
\begin{align*}
\delta_{ij}^t(c_{ij}^t=0) = &\max_{c_{ij}^{t-1}} \big[ D_{ij}^t(c_{ij}^{t-1},c_{ij}^t=0)+s_{ij}^{t-1}(c_{ij}^{t-1})\\
&\quad +\eta_{ij}^{t-1}(c_{ij}^{t-1})+\alpha_{ij}^{t-1}(c_{ij}^{t-1})+\delta_{ij}^{t-1}(c_{ij}^{t-1})\big] \\
=& \max\big[ \eta_{ij}^{t-1}(0)+\alpha_{ij}^{t-1}(0)+\delta_{ij}^{t-1}(0), \\
&\quad -\gamma+s_{ij}^{t-1}+\eta_{ij}^{t-1}(1)+\alpha_{ij}^{t-1}(1)+\delta_{ij}^{t-1}(1)\big], \\
\delta_{ij}^t(c_{ij}^t=1) = &\max_{c_{ij}^{t-1}} \big[ D_{ij}^t(c_{ij}^{t-1},c_{ij}^t=1)+s_{ij}^{t-1}(c_{ij}^{t-1})\\
&\quad+\eta_{ij}^{t-1}(c_{ij}^{t-1})+\alpha_{ij}^{t-1}(c_{ij}^{t-1})+\delta_{ij}^{t-1}(c_{ij}^{t-1})\big] \\
=& \max\big[ -\gamma+\eta_{ij}^{t-1}(0)+\alpha_{ij}^{t-1}(0)+\delta_{ij}^{t-1}(0), \\
&\quad s_{ij}^{t-1}+\eta_{ij}^{t-1}(1)\\
& \quad+\alpha_{ij}^{t-1}(1)+\delta_{ij}^{t-1}(1)\big],
\end{align*}
where $\mathbbm{1}$ denotes the indicator function. After substituting for $\eta_{ij}^{t-1}$ using (\ref{eq_rho})
and assigning $\delta_{ij}^t=\delta_{ij}^t(c_{ij}^t=1)-\delta_{ij}^t(c_{ij}^t=0)$, we obtain the final $\delta$ messages specified in Equation \ref{eq_delta1}.

Similarly, the The $\phi_{ij}^{t-1}$ messages can be similarly derived,
\begin{align*}
\phi_{ij}^{t-1}(c_{ij}^{t-1}=0) = &\max_{c_{ij}^{t}} \big[ D_{ij}^t(c_{ij}^{t-1}=0,c_{ij}^t)+s_{ij}^{t}(c_{ij}^{t})\\
&\quad+\eta_{ij}^{t}(c_{ij}^{t})+\alpha_{ij}^{t}(c_{ij}^{t})+\phi_{ij}^{t}(c_{ij}^{t})\big] \\
=& \max\big[ \eta_{ij}^{t}(0)+\alpha_{ij}^{t}(0)+\phi_{ij}^{t}(0), \\& \quad -\gamma+s_{ij}^{t}+\eta_{ij}^{t}(1)+\alpha_{ij}^{t}(1)+\phi_{ij}^{t}(1)\big] \\
\phi_{ij}^{t-1}(c_{ij}^{t-1}=1) = &\max_{c_{ij}^{t}} \big[ D_{ij}^t(c_{ij}^{t-1}=1,c_{ij}^t)+s_{ij}^{t}(c_{ij}^{t})\\
&\quad+\eta_{ij}^{t}(c_{ij}^{t})+\alpha_{ij}^{t}(c_{ij}^{t})+\phi_{ij}^{t}(c_{ij}^{t})\big], \\
=& \max\big[ -\gamma+\eta_{ij}^{t}(0)+\alpha_{ij}^{t}(0)+\delta_{ij}^{t}(0), \\
&\quad s_{ij}^{t}+\eta_{ij}^{t}(1)\\
&\quad+\alpha_{ij}^{t}(1)+\phi_{ij}^{t}(1)\big].
\end{align*}
After eliminating $\eta_{ij}^{t}$, it is straightforward to show that we obtain the final $\phi$ messages specified in Equation \ref{eq_phi1}.

\section{Experimental Results: Dataset Descriptions}\label{data_descriptions}

\subsection{Gaussian data}\label{data_gaussian}
The initial mixture weights are uniform. At each time step, points in each of the components are 
drawn from the corresponding Gaussian distribution. The first dataset consists of two well-separated 
Gaussians and 40 time steps. The means at the initial time step are set to $[-4,0]$ and $[4,0]$, with 
the covariance of 0.1$I$ ($I$ denotes the identity matrix). At each time step $t$, the first dimension 
of the mean of each component is altered by a random walk with step size 0.1. At $t=19$, the 
covariance matrix is changed to 0.3$I$. The second dataset is generated from two colliding 
Gaussians with the initial means of $[-3,-3]$ and $[3,3]$ and identity covariance. In each of the
time instances $t\!=\!2,\ldots,9$, the mean of the first component is increased by $[0.4,0.4]$. From 
$t=10$ to $t=25$, the means remain constant and the data points are drawn from their respective 
mixture component. The third dataset is generated in a similar way as the second one, with the 
difference that some points change clusters. In particular, at time steps $t=10$ and $t=11$, points 
in the second component switch to the first component with a probability of $0.25$, altering the mixture 
weights. From $t=12$ to $t=25$, the data points maintain the component membership they had at 
$t=11$. Finally, the fourth dataset is generated the same way as the second one for the first $9$ 
time steps. For $t=10$ and $t=11$, data points in the second cluster switch membership with a 
probability of $0.25$ to a new third Gaussian component with mean $[-3,-3]$ and identity covariance.
The data in the synthetic sets was normalized by subtracting the mean across 
time and dividing by the standard deviation of each component. When the components are on the 
same scale, as in the case of synthetic data, normalization is typically not necessary. Nevertheless, 
we perform it for consistency since normalization is generally required for real data sets.

\subsection{Stock prices}\label{data_stock}
Feature vectors were constructed using piecewise normalized derivatives. This feature construction method 
has been shown to yield better clustering results when compared to using raw stock prices or performing 
normalization across all times \cite{gavrilov_2000}. Previously, piecewise normalized derivatives of stock 
market prices were successfully used for evolutionary clustering in \cite{xu_2013}, where NASDAQ stocks 
were clustered using 15-day feature vectors to show the response of an adaptive factor in AFFECT to the 
2008 market crash given a pre-specified number of clusters. The dataset that we study was divided into 
time periods of one month. Within a given month, the difference in closing price between consecutive 
market days was calculated. The final feature vector of normalized derivatives is obtained after normalizing 
the difference vector for each stock to have zero mean and unit standard deviation. Inspection of the plans in each cluster reveals plan type, geographical, and parent company differences between clusters.

\section{Experimental Results: Medicare Star Ratings}
\subsection{Motivation and Background} \label{motivation}
In 2015, \$936 billion were spent on federal health insurance programs, reflecting a \$105 billion increase from the preceding year. Medicare alone, primarily available to those age 65 and over, accounted for \$34 billion of the increase. In 2016, the federal health insurance programs accounted for more than 60\% of the growth in mandatory spending with an increase in spending of \$104 billion \cite{cbo_2016}. The Centers for Medicare and Medicaid Services (CMS) evaluates Medicare plans via a 5-star system, where high star ratings result in bonuses while low star ratings may lead to plan termination. The continually rising cost of healthcare and the rating system that provides financial incentives motivate efforts to identify classes of health plans and study their evolution.

Individuals qualifying for Medicare can choose to either: (a) purchase a Medicare Advantage (MA) plan, in which case the Medicare benefits are provided through a private company approved by Medicare; (b) purchase a non-MA plan such as a Cost plan; or (c) remain with traditional Medicare and pay for services under the Medicare fee-for-service (FFS) structure. Medicare Advantage plans include Health Maintenance Organization (HMO), Preferred Provider Organization (PPO), and Private-Fee-for-Service (PFFS) plans. At a high level, HMO plans cover the members only for medical services received from in-network providers. The member needs to designate a primary care physician (PCP) and receive a referral from the PCP in order to see a specialist. These plans are typically lower cost than PPO plans, in which members do not need to designate a PCP, can see any provider without a referral, and may seek medical care from in-network or out-of-network providers (if allowed, the latter typically incurs higher costs). Members of PFFS plans pay for some of the highest costs in exchange for flexibility -- they can see any provider that agrees to the plan's terms and conditions. 

To form an overall star rating for a Medicare plan, CMS evaluates various aspects of the plan including measures encompassing health screenings and tests, management of chronic conditions, hospital readmission rate, customer service, and member experience. The overall star rating for the plan is a weighted average of the star ratings for the individual measures. The ratings both provide accountability for the health plans and help consumers select a plan. Studies on the effect of the star ratings on consumer choice have yielded inconclusive results. One study found the 2009 ratings deterred people from choosing low-rated plans, with the effect disappearing in 2010; therefore, ratings did not have an effect on enrollees' plan choice \cite{darden_2015}. However, another study using 2011 star ratings found a positive association between the ratings and enrollment, e.g., a 1-star rating increase was associated with a higher likelihood to enroll \cite{reid_2013}. Five-star plans are further awarded with a special open enrollment period 11 months longer than other plans. In addition to potential enrollment benefits, plans have financial motivation to achieve higher ratings. In particular, in 2012 CMS began awarding bonuses to plans rated at 4 or more stars while plans that receive fewer than 3 stars for 3 years in a row may be discontinued by CMS. Note that the thresholds for assigning stars to individual measures are calculated based on hierarchical clustering of the raw data (e.g., a medication adherence score from 0 to 100) and simple statistics such as deviations from the mean. Plans may seek to improve their performance measured by the raw data since a plan that has a stagnant raw score may see its star rating decrease if other plans are improving.

Recent studies have shown that higher adherence to diabetes \cite{stuart_2011,roebuck_2011}, cholesterol \cite{pittman_2011,roebuck_2011}, and hypertension \cite{pittman_2010,roebuck_2011} medications leads to lower overall healthcare spending despite higher drug spending. The reduction in overall spending is largely due to fewer hospitalizations and emergency department visits for cardiovascular disease \cite{pittman_2010,pittman_2011,roebuck_2011}. Roebuck et al. studied medication adherence in patients with congestive heart failure (CHF), diabetes, hypertension, and dyslipidemia (i.e. elevated cholesterol and/or tryglicerides) \cite{roebuck_2011}. They found adherent patients ages 65 and over had on average between 1.88 (for dyslipidemia) and 5.87 (for CHF) fewer inpatient hospital days a year. After accounting for increased drug costs, these patients also had on average yearly healthcare savings of between \$1857 (for dyslipidemia) and \$7893 (for CHF), with average savings of over \$5000 for diabetes and hypertension.


We seek to understand the dynamics of medication adherence by grouping MA plans using evolutionary affinity propagation (EAP) and characterizing the resulting clusters.





\subsection{Data and Algorithm Implementation}
We consider diabetes, hypertension, and cholesterol medication adherence data from MA plans with a prescription drug plan in 2012-2016 \cite{cmsdata}. The data is available as both the raw score (0-100) and the star rating assigned to each individual measure. The medication adherence raw score indicates the percentage of adherent members in the measurement period, e.g., a year. A plan member is considered adherent for a specific class of medications if the amount of that class of medication filled would cover 80\% of the days in the measurement period, and the members with measured adherence are those ages 18 and older with at least two fills of medications in the drug classes of interest.

To be included in the study, the raw data had to be available for at least 2 years between 2012 and 2016, with at least one score available for each of the 3 categories of medication adherence. The medication adherence data is reported on a yearly basis ($T=5$). Out of 1018 MA plans, 318 did not have any medication adherence data available in any year, 31 did not have data for one or two medication adherence measures in any year, and 72 had medication adherence data available for only one year. The final dataset consisted of 597 MA plans from 45 states,  the District of Columbia, and Puerto Rico. 

Three-dimensional feature vectors collecting medication adherence raw scores were constructed for each plan and each time point. Plans were considered to be active if they were associated with any star rating measures in a given year. Missing values for time steps where a plan was active were imputed using the previous known value or the first known value. The similarity between plans was defined as the negative squared Euclidean distance evaluated on normalized imputed data, where data were normalized by subtracting the global mean and dividing by the global standard deviation for each feature. Note that the global normalization is necessary in order to preserve the known overall upward trend in medication adherence scores over time; a global preference aids in determining if this upward trend affects the resulting number of clusters. EAP was implemented with $\omega=1,\gamma=2$, a minimum cluster size of $20$ for consensus node creation, AP damping parameter $\mu=0.5$, a maximum of 500 iterations, 20 iterations without changes in the exemplar set for convergence, and the preference set to the minimum similarity over all time steps. The data were also clustered with AP and EAP without consensus nodes, with the same parameters as EAP where applicable, and with AFFECT using the modularity criterion to determine the number of clusters.

\subsection{Clustering Analysis}
\begin{figure*}[!h]
\begin{centering}
\includegraphics[width=6in]{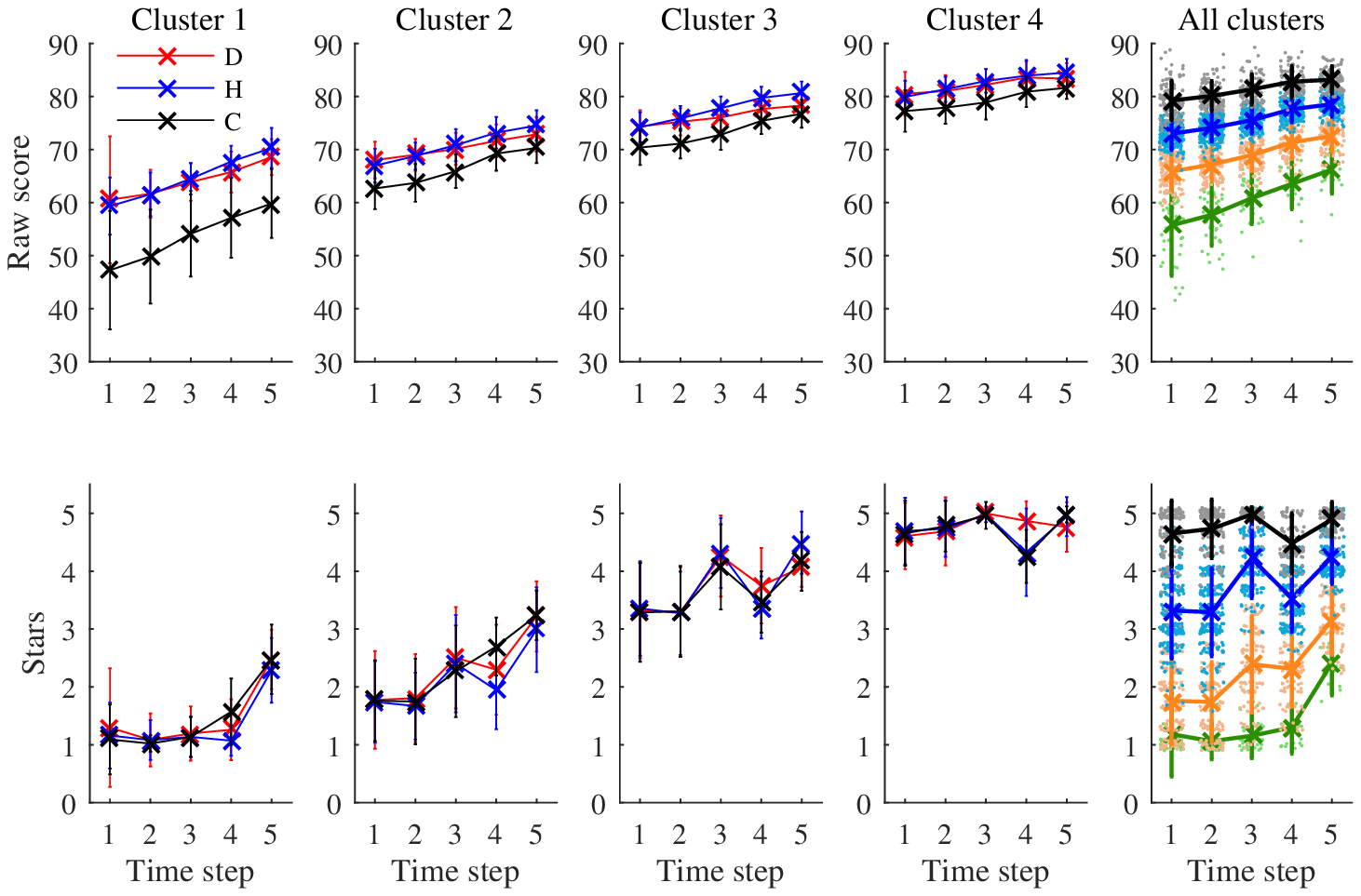}
\caption{Average non-imputed raw scores (top panel) and star rating (bottom panel) for diabetes (D, red), hypertension (H, blue) and cholesterol (C, black) medication adherence per cluster for years 2012-2016. The right panel compares the average medication adherence score for each plan (jittered points) in each cluster (color). The average for clusters 1, 2, 3, and 4 are shown by the green, orange, blue, and black lines. Error bars represent standard deviation.}
\label{fig_dataStars}
\end{centering}
\end{figure*}

EAP revealed 4 main clusters that track throughout all 5 time steps. We refer to the clusters by numbers -- cluster 1, 2, 3, and 4 -- ranking the average raw scores of the cluster members from worst to best (Figure \ref{fig_dataStars}). Table \ref{table_clustSize} details the size of the clusters at each time step. 

\begin{table}
\renewcommand{\arraystretch}{1.3}
\caption{Cluster Size by Year}
\label{table_clustSize}
\centering
\begin{center}
\begin{tabular}{cccccc}
\hline
 & 2012 & 2013 & 2014 & 2015 & 2016 \\
\hline
Cluster 1 &  44 & 52 & 40 & 29 & 22\\
\hline
Cluster 2 & 146 & 130 & 122 & 109 & 93\\
\hline
Cluster 3 & 231 & 245 & 204 & 188 & 154\\
\hline
Cluster 4 & 137 & 143 & 110 & 111 & 105\\
\hline
All plans & 558 & 570 & 476 & 437 & 374\\
\hline
\end{tabular}
\end{center}
\end{table}

Most plans ($90-95\%$) remain in the same cluster across consecutive time steps. Plans that do change cluster membership typically switch between adjacent clusters (e.g., from cluster 3 to clusters 2 or 4), with the exception of 3 plans. One plan, Cigna-HealthSpring in Florida, jumped from cluster 1 to cluster 4. This is a result of a large increase in the raw measures (diabetes, hypertension, cholesterol) between 2013 and 2014 (75.8, 74.5, 54.5) to (87, 85, 73) and a subsequent increase in individual star ratings from (4, 3, 1) to (5, 5, 4). Two other Florida plans jumped from cluster 2 to cluster 4. These plans were associated with similar increases in raw scores, (15.7, 12.2, 12.9) and (8, 7, 11), and star ratings.

Clustering with AP, EAP:noCN, and AFFECT mostly yielded both a larger number of clusters and much more frequent cluster membership changes. Table \ref{tableS_clustNum} displays the number of clusters found at each time step for each algorithm. 

\begin{table}
\renewcommand{\arraystretch}{1.3}
\caption{Number of Clusters per Year}
\label{tableS_clustNum}
\centering
\begin{center}
\begin{tabular}{cccccc}
\hline
 & 2012 & 2013 & 2014 & 2015 & 2016 \\
\hline
EAP &  4 & 4 & 4 & 4 & 4\\
\hline
EAP:noCN & 3 & 4 & 2 & 2 & 2\\
\hline
AP & 4 & 98 & 203 & 3 & 292\\
\hline
AFFECT & 10 & 10 & 9 & 10 & 10\\
\hline
\end{tabular}
\end{center}
\end{table}

\begin{figure}[!h]
\begin{centering}
\includegraphics[width=3.5in]{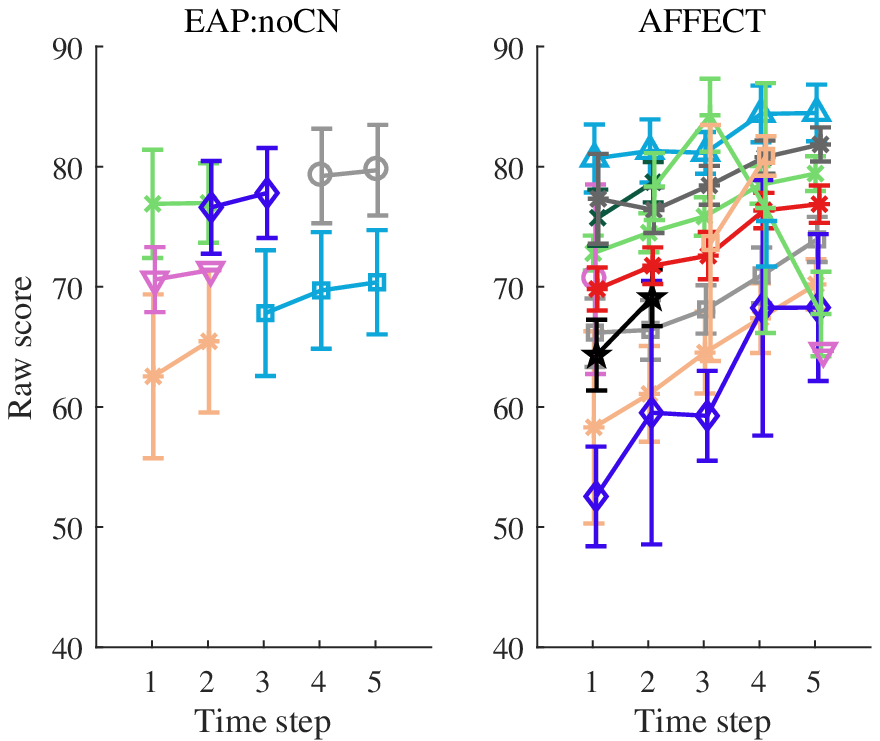}
\caption{Average of non-imputed medication adherence scores for each cluster. Error bars represent standard deviation.}
\label{fig_stars_base1}
\end{centering}
\end{figure}

The lower number of clusters found by EAP:noCN does not indicate stability in the clustering, since up to 72\% of the plans may change cluster membership at consecutive time steps. With AFFECT, up to 51\% of plans change cluster membership at consecutive time steps. Both of these solutions have clusters that only exist at one or two time steps (Fig \ref{fig_stars_base1})). Additionally, the clusters formed by AFFECT are largely overlapping (Fig \ref{fig_stars_base1}) even when plotted by the individual medication components (not shown). These results, along with the more detailed analysis in the subsequent sections, indicate EAP is capable of identifying stable clusters and tracking these clusters over time when other evolutionary clustering algorithms may find less informative clusters.

All clusters in EAP have an upward trend in medication adherence scores, suggesting that inclusion of such a measure in the CMS star ratings is having the intended effect. The average adherence score of cluster 2 in 2016 ($t=5$) is similar to that of cluster 3 in 2012 ($t=1$); a similar relationship exists between cluster 3 and cluster 4. The average cholesterol score appears to be decisive for membership in cluster 1, which gathers plans characterized by a cholesterol adherence score that is much worse than the other two medication adherence measures. From the bottom panel of Figure \ref{fig_dataStars}, we see that an overall improvement in the medication adherence raw scores does not necessarily indicate an increase in star ratings; this is due to the specifics of determining star ratings (see Section \ref{motivation} for details).

\begin{figure*}[h!]
\begin{centering}
\includegraphics[width=5.7in]{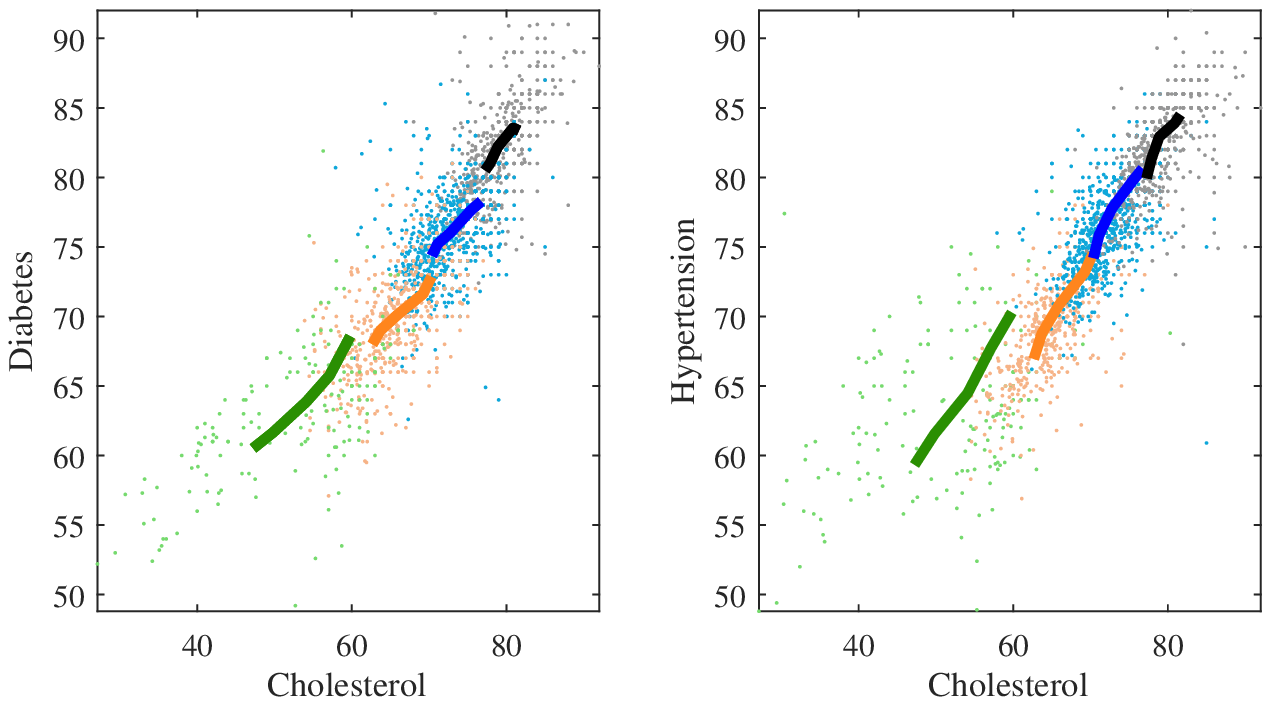}
\caption{Average paths for diabetes vs. cholesterol (left) and hypertension vs. cholesterol (right) plots for clusters 1 (green), 2 (orange), 3 (blue), and 4 (black). The lines connect the average values across time, moving in the direction of the upper right corner. The points indicate values from all plans in the cluster at all times.}
\label{fig_vsChol}
\end{centering}
\end{figure*}

The path plots in Figure \ref{fig_vsChol} show how the relationship between two variables evolves over time. Specifically, we plot the mean medication adherence 
values for each cluster and connect those values across time. The bottom left corner of each line represents the means at $t=1$ and the upper right corner represents the means at $t=5$; as noted earlier, medication adherence scores on average improve over time. A closer examination of such plots reveals that the average paths for diabetes vs. cholesterol 
are linearly separable. Average paths for hypertension vs. cholesterol tend to overlap slightly at the end points (e.g., cluster 2 at $t=5$ with cluster 3 at $t=1$). 
Figure \ref{fig_vsChol} emphasizes the idea that the clusters identified by EAP reflect different levels of medication adherence, with measures improving over time 
for all clusters. We can also see that cluster 1 has the greatest improvement in medication adherence raw scores while cluster 4 has the least improvement across time. The large improvement in cluster 1 is likely due to a combination of a survival of the fittest effect, where some the poorly performing plans disappear, and a greater overall improvement in adherence measures than other clusters. The smaller improvement in cluster 4 suggests there is a ceiling on a realistic maximal value of 
medication adherence.

Inspection of the plans in each cluster reveals plan type, geographical, and parent company differences between clusters.

\subsubsection{Plan types and medication adherence}
A natural question that arises in the analysis is whether medication adherence scores vary between different plan types. To this
end, we compare the scores of HMO, PPO, Cost, and PFFS plans. HMO plans are prevalent in all clusters, which comes as no surprise since they are the most common type of a plan. Closer inspection reveals that they are more dominant in clusters 1 and 2, where they make up $72-88\%$ of each cluster at different times, than in clusters 3 and 4, where their representation is $47-69\%$ of the cluster population. PPO plans, on the other hand, are present primarily in clusters 3 and 4, with cluster 4 having the highest composition of PPO plans ($29-34\%$ of the cluster members). All but one of the Cost plans are grouped in cluster 4, with the remaining plan being part of cluster 3. Finally, PFFS plans are grouped into clusters 2, 3, and 4, with most of the plans in cluster 3. These results suggest that plans which provide members with more flexibility tend to perform better in the medication adherence measures. An important consideration, however, is that plans that allow for more flexibility at a higher cost may also have a member population that is more likely to be adherent to medications. Analysis of plan member demographics, which are not available to us, would be necessary to determine relationships between plan type, member population, and medication adherence.

\subsubsection{States and medication adherence} \label{sec_states}

\begin{figure*}[!h]
\begin{centering}
\includegraphics[width=5.7in]{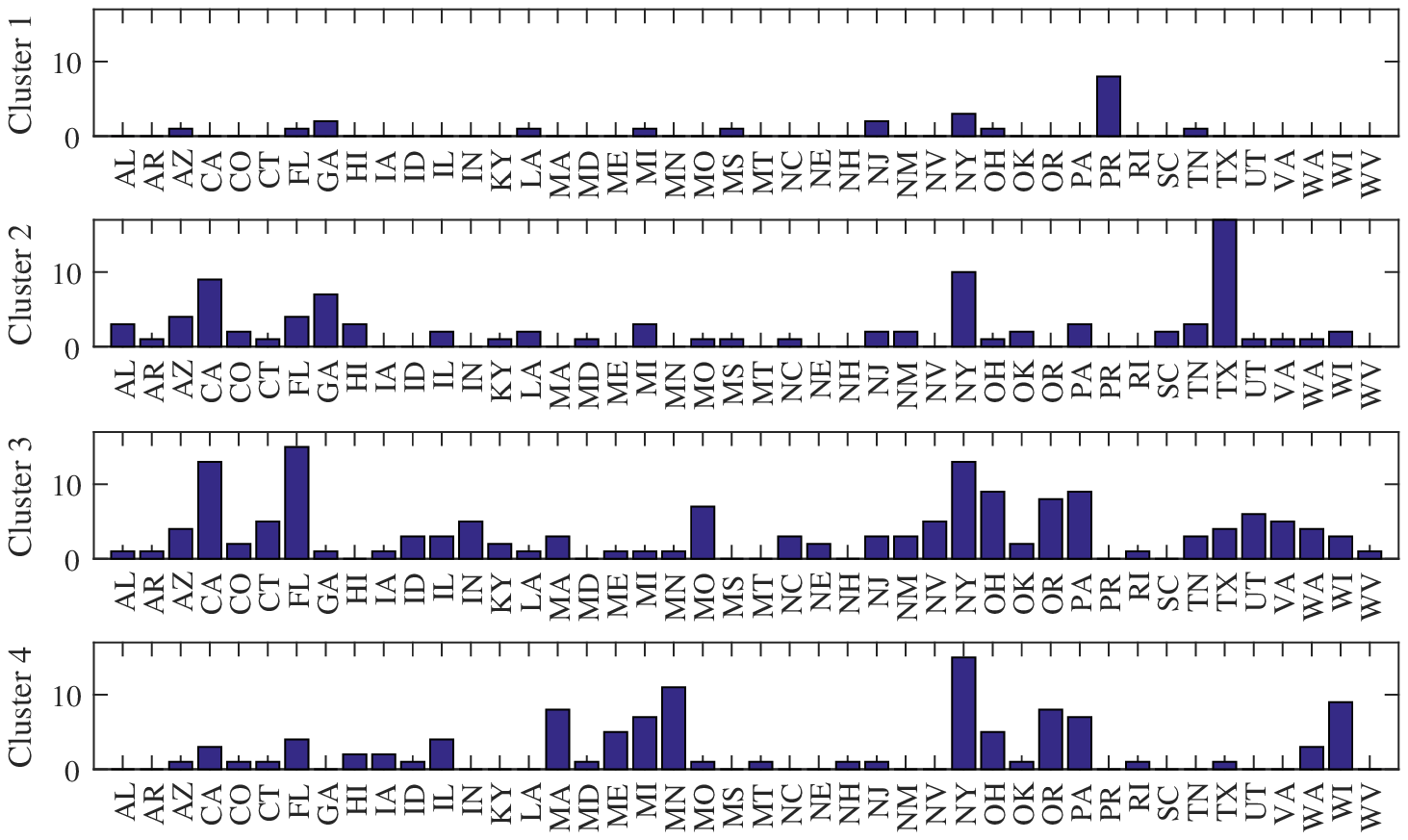}
\caption{Number of plans from each state in each cluster in 2016.}
\label{fig_stateAbs}
\end{centering}
\end{figure*}

\begin{figure*}[!h]
\begin{centering}
\includegraphics[width=5.7in]{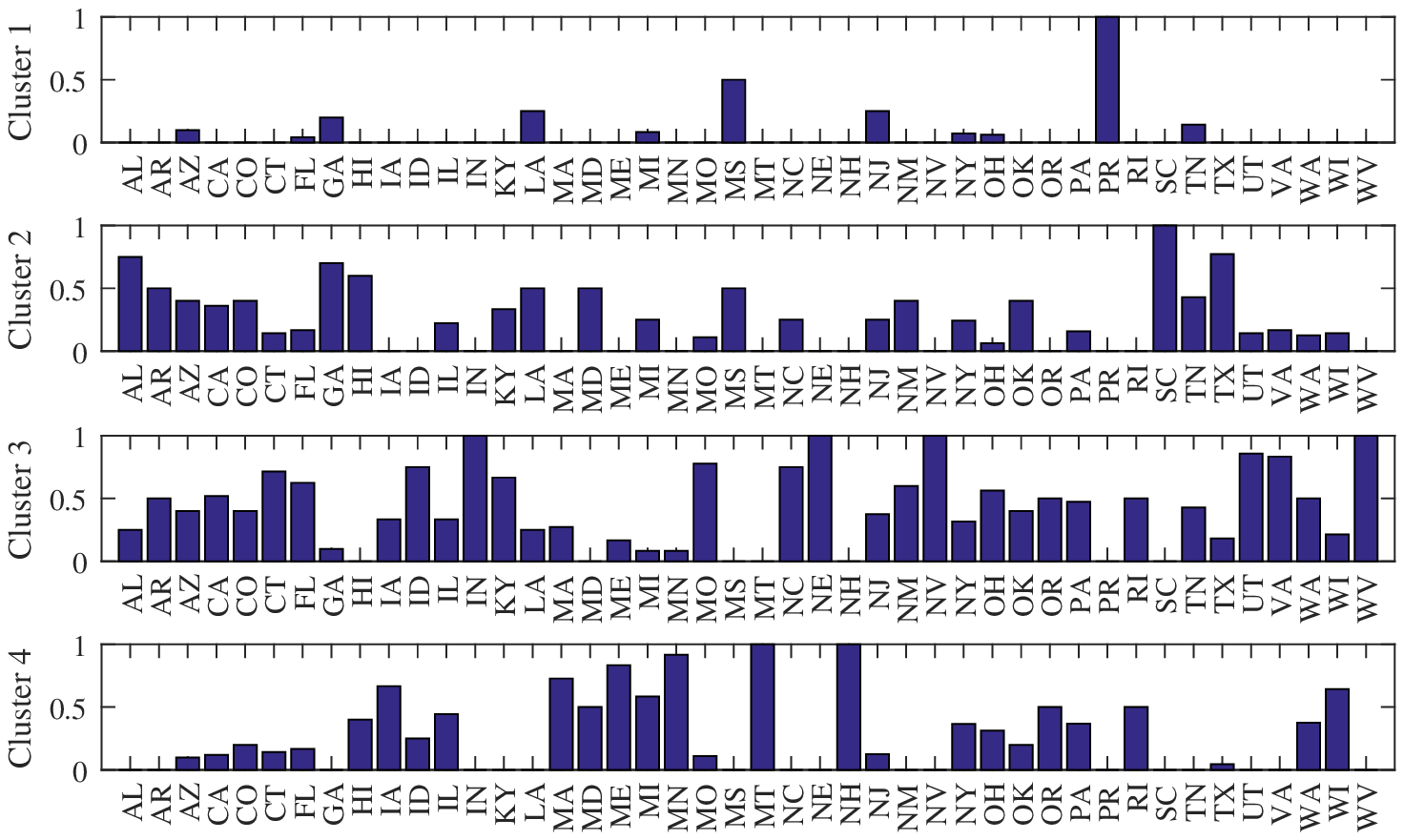}
\caption{Fraction of plans from each state in each cluster in 2016.}
\label{fig_statePerc}
\end{centering}
\end{figure*}

Some of the discovered clusters are dominated by a group of states or territories, suggesting geographic variations in medication adherence. Figures \ref{fig_stateAbs} and \ref{fig_statePerc} show the number and percentage of plans from each state in each cluster in 2016 ($t=5$), respectively.

Cluster 1, the worst performing one, is dominated by plans from Puerto Rico, all of which cluster together. A letter from CMS on November 2015, regarding star ratings for 2017 and available at \cite{cmsdata}, states that Medicare in Puerto Rico suffers unique implementation challenges. In particular, despite many low-income individuals participating in Medicare in Puerto Rico, the Medicare Part D low income subsidy is not available. This, at least in part, may explain the poor performance of Puerto Rico plans in terms of medication adherence, leading to their grouping in cluster 1. CMS addressed the issue by not having medication adherence measures count towards the overall star rating of Puerto Rico plans in future, though they will still be used to calculate improvement metrics that contribute to the overall rating. 

Cluster 2 is dominated by Texas plans. Cluster 3 is dominated by Florida, California, New York, Ohio, Pennsylvania, Missouri, and Oregon plans. Several Florida plans move from cluster 2 to cluster 3 as time passes, leading to a large presence in cluster 3 at the last time step as shown in the figures. Cluster 4 is dominated by New York, Wisconsin, Montana, Massachusetts, Oregon, Pennsylvania, and Michigan. Most of New York plans are divided between clusters 2, 3, and 4, though generally more plans are grouped in clusters 3 and 4 than in cluster 2. Previous studies have found Massachusetts and California to have the best Medicare Advantage plans in a by-state analysis \cite{soria-saucedo_2016}, and New England \cite{couto_2014} to have the best medication adherence in the country in a regional analysis. In our analysis, most of the California plans are in cluster 3, though a few are in cluster 4 at all times. New England states have a strong preference towards the best clusters, with New Hampshire plans all grouping in cluster 4, Massachusetts and Maine plans mostly falling in cluster 4, and Connecticut and Rhode Island plans primarily grouping in cluster 3. No plans from Vermont met our data inclusion criteria.

\subsubsection{Parent companies and medication adherence}

Analysis of the health plan parent companies reveals that some parent companies tend to group together while others are more diverse and spread across clusters. Examination of parent companies that dominate each cluster reveals Kaiser plans ($n=6$) are grouped in cluster 4, with one plan belonging to cluster 3 in 2015 and 2016. Other parent companies with significant representation in cluster 4 include UnitedHealth Group, Humana, and Aetna. However, Humana and Aetna plans are primarily grouped in cluster 3, while UnitedHealth Group plans are mostly split between clusters 2 and 3. WellPoint, which later changed its name to Anthem, was also primarily grouped in cluster 3. Cluster 1 did not have any parent company with more than 5 plans. 

We additionally sought to determine if the plans under the same parent company tend to group together. For this analysis, we only considered parent companies with more than one plan at a given time step and calculated the percentage of plans in the cluster with parent companies whose plans all cluster together. Cluster 4 has the highest such percentage, $26-33\%$ of the plans and $54-65\%$ of the parent companies. A similar analysis of clusters 2 and 3 reveals that only $12-21\%$ of the plans therein belong to parent companies that completely clustered together ($44-60\%$ of the companies). This suggests that some smaller parent companies with plans that perform well do so uniformly across all plans.

\subsection{Summary}

We demonstrated that an application of evolutionary affinity propagation to medication adherence data yields groups of plans that behave similarly and are interpretable. In particular, for the given data set we identified four clusters that can be interpreted as containing plans characterized by different levels of medication adherence. Detailed analysis of cluster composition reveals differences in state, plan type, and parent company distribution between clusters, suggesting these factors are related to medication adherence.

\bibliographystyle{IEEEtran}


\begin{thebibliography}{}
\providecommand{\url}[1]{#1}
\csname url@samestyle\endcsname
\providecommand{\newblock}{\relax}
\providecommand{\bibinfo}[2]{#2}
\providecommand{\BIBentrySTDinterwordspacing}{\spaceskip=0pt\relax}
\providecommand{\BIBentryALTinterwordstretchfactor}{4}
\providecommand{\BIBentryALTinterwordspacing}{\spaceskip=\fontdimen2\font plus
\BIBentryALTinterwordstretchfactor\fontdimen3\font minus
  \fontdimen4\font\relax}
\providecommand{\BIBforeignlanguage}[2]{{%
\expandafter\ifx\csname l@#1\endcsname\relax
\typeout{** WARNING: IEEEtran.bst: No hyphenation pattern has been}%
\typeout{** loaded for the language `#1'. Using the pattern for}%
\typeout{** the default language instead.}%
\else
\language=\csname l@#1\endcsname
\fi
#2}}
\providecommand{\BIBdecl}{\relax}
\BIBdecl

\end{thebibliography}


\begin{thebibliography}{10}
\providecommand{\url}[1]{#1}
\csname url@samestyle\endcsname
\providecommand{\newblock}{\relax}
\providecommand{\bibinfo}[2]{#2}
\providecommand{\BIBentrySTDinterwordspacing}{\spaceskip=0pt\relax}
\providecommand{\BIBentryALTinterwordstretchfactor}{4}
\providecommand{\BIBentryALTinterwordspacing}{\spaceskip=\fontdimen2\font plus
\BIBentryALTinterwordstretchfactor\fontdimen3\font minus
  \fontdimen4\font\relax}
\providecommand{\BIBforeignlanguage}[2]{{%
\expandafter\ifx\csname l@#1\endcsname\relax
\typeout{** WARNING: IEEEtran.bst: No hyphenation pattern has been}%
\typeout{** loaded for the language `#1'. Using the pattern for}%
\typeout{** the default language instead.}%
\else
\language=\csname l@#1\endcsname
\fi
#2}}
\providecommand{\BIBdecl}{\relax}
\BIBdecl

\bibitem{tant07}
C.~Tantipathananandh, T.~Berger-Wolf, and D.~Kempe, ``A framework for community
  identification in dynamic social networks,'' in \emph{Proceedings of the 13th
  ACM SIGKDD international conference on Knowledge discovery and data
  mining}.\hskip 1em plus 0.5em minus 0.4em\relax ACM, 2007, pp. 717--726.

\bibitem{Li:2004aa}
Y.~Li, J.~Han, and J.~Yang, ``Clustering moving objects,'' in \emph{Proceedings
  of the tenth ACM SIGKDD international conference on Knowledge discovery and
  data mining}, 2004, pp. 617--622.

\bibitem{Fenn:2009aa}
D.~J. Fenn, M.~A. Porter, M.~McDonald, S.~Williams, N.~F. Johnson, and N.~S.
  Jones, ``Dynamic communities in multichannel data: {A}n application to the
  foreign exchange market during the 2007--2008 credit crisis,'' \emph{Chaos:
  An interdisciplinary journal of nonlinear science}, vol.~19, no.~3, p.
  033119, 2009.

\bibitem{chakrabarti_2006}
D.~Chakrabarti, R.~Kumar, and A.~Tomkins, ``Evolutionary clustering,'' in
  \emph{Proceedings of the 12th {ACM} {SIGKDD} {International} {Conf.} on
  {Knowledge} {Discovery} and {Data} {Mining}}.\hskip 1em plus 0.5em minus
  0.4em\relax New York, NY, USA: ACM, 2006, pp. 554--560.

\bibitem{ahmed_2008}
A.~Ahmed and E.~Xing, ``Dynamic non-parametric mixture models and the recurrent
  chinese restaurant process: with applications to evolutionary clustering,''
  in \emph{Proceedings of the 2008 {SIAM} {International} {Conference} on
  {Data} {Mining}}.\hskip 1em plus 0.5em minus 0.4em\relax Society for
  Industrial and Applied Mathematics, Apr. 2008, pp. 219--230.

\bibitem{xu_2013}
K.~S. Xu, M.~Kliger, and A.~O. Hero~III, ``Adaptive evolutionary clustering,''
  \emph{Data Mining and Knowledge Discovery}, vol.~28, no.~2, pp. 304--336,
  Jan. 2013.

\bibitem{chi_2007}
Y.~Chi, X.~Song, D.~Zhou, K.~Hino, and B.~L. Tseng, ``Evolutionary spectral
  clustering by incorporating temporal smoothness,'' in \emph{Proceedings of
  the 13th {ACM} {SIGKDD} {International} {Conference} on {Knowledge}
  {Discovery} and {Data} {Mining}}.\hskip 1em plus 0.5em minus 0.4em\relax New
  York, NY, USA: ACM, 2007, pp. 153--162.

\bibitem{chi_2009}
------, ``On evolutionary spectral clustering,'' \emph{ACM Trans. Knowl.
  Discov. Data}, vol.~3, no.~4, pp. 17:1--17:30, Dec. 2009.

\bibitem{xu_2010}
K.~S. Xu, M.~Kliger, and A.~O. Hero~III, ``Tracking communities of spammers by
  evolutionary clustering,'' in \emph{International {Conference} on {Machine}
  {Learning} {Workshop} on {Social} {Analytics}: {Learning} from {Human}
  {Interactions}}, 2010.

\bibitem{czink_2007}
N.~Czink, R.~Tian, S.~Wyne, F.~Tufvesson, J.-P. Nuutinen, J.~Ylitalo, E.~Bonek,
  and A.~Molisch, ``Tracking time-variant cluster parameters in {MIMO} channel
  measurements,'' in \emph{Second {International} {Conference} on
  {Communications} and {Networking} in {China}}, Aug. 2007, pp. 1147--1151.

\bibitem{ahmed_2010}
E.~P.~X. Amr~Ahmed, ``Timeline: A dynamic hierarchical dirichlet process model
  for recovering birth/death and evolution of topics in text stream,'' in
  \emph{Proc. of the 26th {Conf.} on {Uncertainty} in {Artificial} {Intell.}},
  2010, pp. 20--29.

\bibitem{xu_2008}
T.~Xu, Z.~Zhang, P.~Yu, and B.~Long, ``Evolutionary clustering by hierarchical
  dirichlet process with hidden markov state,'' in \emph{Eighth {IEEE}
  {International} {Conference} on {Data} {Mining}, 2008. {ICDM} '08}, Dec.
  2008, pp. 658--667.

\bibitem{folino_2014}
F.~Folino and C.~Pizzuti, ``An evolutionary multiobjective approach for
  community discovery in dynamic networks,'' \emph{IEEE Transactions on
  Knowledge and Data Engineering}, vol.~26, no.~8, pp. 1838--1852, Aug. 2014.

\bibitem{arze17}
N.~M. Arzeno and H.~Vikalo, ``Evolutionary affinity propagation,'' in
  \emph{IEEE Conference on Acoustic, Speech and Signal Processing}, 2017.

\bibitem{greene_2010}
D.~Greene, D.~Doyle, and P.~Cunningham, ``Tracking the evolution of communities
  in dynamic social networks,'' in \emph{2010 International Conference on
  Advances in Social Networks Analysis and Mining}, Aug. 2010, pp. 176--183.

\bibitem{brodka_2012}
P.~Br{\'o}dka, S.~Saganowski, and P.~Kazienko, ``{GED}: the method for group
  evolution discovery in social networks,'' Jul. 2012.

\bibitem{gunnemann_2011}
S.~G{\"u}nnemann, H.~Kremer, C.~Laufk{\"o}tter, and T.~Seidl,
  ``\BIBforeignlanguage{en}{Tracing evolving subspace clusters in temporal
  climate data},'' \emph{\BIBforeignlanguage{en}{Data Mining and Knowledge
  Discovery}}, vol.~24, no.~2, pp. 387--410, Sep. 2011.

\bibitem{kim_2015}
Y.-M. Kim, J.~Velcin, S.~Bonnevay, and M.-A. Rizoiu,
  ``\BIBforeignlanguage{en}{Temporal multinomial mixture for instance-oriented
  evolutionary clustering},'' in \emph{\BIBforeignlanguage{en}{Advances in
  {Information} {Retrieval}}}, Mar. 2015, pp. 593--604.

\bibitem{yang_2010}
T.~Yang, Y.~Chi, S.~Zhu, Y.~Gong, and R.~Jin,
  ``\BIBforeignlanguage{en}{Detecting communities and their evolutions in
  dynamic social networks---a {Bayesian} approach},''
  \emph{\BIBforeignlanguage{en}{Machine Learning}}, vol.~82, no.~2, pp.
  157--189, Sep. 2010.

\bibitem{fortunato_2010}
S.~Fortunato, ``Community detection in graphs,'' \emph{Physics Reports}, vol.
  486, no. 3--5, pp. 75--174, Feb. 2010.

\bibitem{jia_2014}
X.~Jia, N.~Du, J.~Gao, and A.~Zhang, ``Analysis on community variational trend
  in dynamic networks,'' in \emph{Proceedings of the 23rd {ACM} {International}
  {Conference} on {Conference} on {Information} and {Knowledge}
  {Management}}.\hskip 1em plus 0.5em minus 0.4em\relax New York, NY, USA: ACM,
  2014, pp. 151--160.

\bibitem{frey_2007}
B.~J. Frey and D.~Dueck, ``\BIBforeignlanguage{en}{Clustering by passing
  messages between data points},'' \emph{\BIBforeignlanguage{en}{Science}},
  vol. 315, no. 5814, pp. 972--976, Feb. 2007.

\bibitem{givoni_2009}
I.~E. Givoni and B.~J. Frey, ``A binary variable model for affinity
  propagation,'' \emph{Neural computation}, vol.~21, no.~6, pp. 1589--1600,
  Jun. 2009.

\bibitem{givoni_2009ss}
------, ``Semi-supervised affinity propagation with instance-level
  constraints,'' in \emph{Proceedings of the Twelfth International Conference
  on Artificial Intelligence and Statistics}, 2009, pp. 161--168.

\bibitem{leone_2008}
M.~Leone, Sumedha, and M.~Weigt, ``Unsupervised and semi-supervised clustering
  by message passing: soft-constraint affinity propagation,'' \emph{The
  European Physical Journal B}, vol.~66, no.~1, pp. 125--135, Oct. 2008.

\bibitem{arzeno_2015}
N.~M. Arzeno and H.~Vikalo, ``\BIBforeignlanguage{eng}{Semi-supervised affinity
  propagation with soft instance-level constraints},''
  \emph{\BIBforeignlanguage{eng}{IEEE {Transactions} on {Pattern} {Analysis}
  and {Machine} {Intelligence}}}, vol.~37, no.~5, pp. 1041--1052, May 2015.

\bibitem{givoni_2012}
I.~Givoni, C.~Chung, and B.~J. Frey, ``Hierarchical affinity propagation,''
  {arXiv} e-print 1202.3722, Feb. 2012.

\bibitem{wang_2013}
C.-D. Wang, J.-H. Lai, C.~Y. Suen, and J.-Y. Zhu, ``Multi-exemplar affinity
  propagation,'' \emph{IEEE {Transactions} on {Pattern} {Analysis} and
  {Machine} {Intelligence}}, vol.~35, no.~9, pp. 2223 -- 2237, 2013.

\bibitem{fujiwara_2015}
Y.~Fujiwara, M.~Nakatsuji, H.~Shiokawa, Y.~Ida, and M.~Toyoda, ``Adaptive
  message update for fast affinity propagation,'' in \emph{Proc. of the 21th
  {ACM} {SIGKDD} {Intern.} {Conf.} on {Knowledge} {Discovery} and {Data}
  {Mining}}.\hskip 1em plus 0.5em minus 0.4em\relax New York, NY, USA: ACM,
  2015, pp. 309--318.

\bibitem{quera_2013}
V.~Quera, F.~S. Beltran, I.~Givoni, and R.~Dolado, ``Determining shoal
  membership using affinity propagation,'' \emph{Behavioural Brain Res.}, vol.
  241, pp. 38--49, Mar. 2013.

\bibitem{bishop_2006}
C.~M. Bishop, \emph{\BIBforeignlanguage{en}{Pattern Recognition and Machine
  Learning}}.\hskip 1em plus 0.5em minus 0.4em\relax Springer, Aug. 2006.

\bibitem{sun_2014}
L.~Sun and C.~Guo, ``Incremental affinity propagation clustering based on
  message passing,'' \emph{IEEE Transactions on Knowledge and Data
  Engineering}, vol.~26, no.~11, pp. 2731--2744, 2014.

\bibitem{rosswog_2008}
J.~Rosswog and K.~Ghose, ``Detecting and tracking spatio-temporal clusters with
  adaptive history filtering,'' in \emph{{IEEE} International Conference on
  Data Mining Workshops}, Dec. 2008, pp. 448--457.

\bibitem{newman_2006}
M.~E.~J. Newman, ``Modularity and community structure in networks,''
  \emph{Proceedings of the National Academy of Sciences}, vol. 103, no.~23, pp.
  8577--8582, Jun. 2006.

\bibitem{rousseeuw_1987}
P.~J. Rousseeuw, ``Silhouettes: {A} graphical aid to the interpretation and
  validation of cluster analysis,'' \emph{Journal of Computational and Applied
  Mathematics}, vol.~20, pp. 53--65, Nov. 1987.

\bibitem{rand_1971}
W.~M. Rand, ``Objective criteria for the evaluation of clustering methods,''
  \emph{Journal of the American Statistical Association}, vol.~66, no. 336, pp.
  846--850, Dec. 1971.

\bibitem{oliver_2004}
M.~J. Oliver, S.~Glenn, J.~T. Kohut, A.~J. Irwin, O.~M. Schofield, M.~A.
  Moline, and W.~P. Bissett, ``\BIBforeignlanguage{en}{Bioinformatic approaches
  for objective detection of water masses on continental shelves},''
  \emph{\BIBforeignlanguage{en}{Journal of Geophysical Research: Oceans}}, vol.
  109, no.~C7, p. C07S04, Jul. 2004.

\bibitem{chen_2011}
S.~Chen, Y.~Li, J.~Hu, A.~Zheng, L.~Huang, and Y.~Lin, ``Multiparameter cluster
  analysis of seasonal variation of water masses in the eastern {Beibu}
  {Gulf},'' \emph{Journal of Oceanography}, vol.~67, no.~6, pp. 709--718, Oct.
  2011.

\bibitem{qi_2014}
J.~Qi, B.~Yin, Q.~Zhang, D.~Yang, and Z.~Xu, ``\BIBforeignlanguage{en}{Analysis
  of seasonal variation of water masses in {East} {China} {Sea}},''
  \emph{\BIBforeignlanguage{en}{Chinese Journal of Oceanology and Limnology}},
  vol.~32, no.~4, pp. 958--971, Jul. 2014.

\bibitem{roemmich_2009}
D.~Roemmich and J.~Gilson, ``The 2004--2008 mean and annual cycle of
  temperature, salinity, and steric height in the global ocean from the {Argo}
  {Program},'' \emph{Progress in Oceanography}, vol.~82, no.~2, pp. 81--100,
  Aug. 2009.

\bibitem{talley_indian}
L.~D. Talley, G.~L. Pickard, W.~J. Emery, and J.~H. Swift, ``Chapter 11 -
  {Indian} {Ocean},'' in \emph{Descriptive {Physical} {Oceanography} ({Sixth}
  {Edition})}.\hskip 1em plus 0.5em minus 0.4em\relax Boston: Academic Press,
  2011, pp. 363--399.

\bibitem{rusciano_2012}
E.~Rusciano, S.~Speich, and M.~Ollitrault, ``\BIBforeignlanguage{en}{Interocean
  exchanges and the spreading of {Antarctic} {Intermediate} {Water} south of
  {Africa}},'' \emph{\BIBforeignlanguage{en}{Journal of Geophysical Research:
  Oceans}}, vol. 117, no. C10, p. C10010, Oct. 2012.

\bibitem{talley_southern}
L.~D. Talley, G.~L. Pickard, W.~J. Emery, and J.~H. Swift, ``Chapter 13 -
  {Southern} {Ocean},'' in \emph{Descriptive {Physical} {Oceanography} ({Sixth}
  {Edition})}.\hskip 1em plus 0.5em minus 0.4em\relax Boston: Academic Press,
  2011, pp. 437--471.

\bibitem{talley_atlantic}
------, ``Chapter 9 - {Atlantic} {Ocean},'' in \emph{Descriptive {Physical}
  {Oceanography} ({Sixth} {Edition})}.\hskip 1em plus 0.5em minus 0.4em\relax
  Boston: Academic Press, 2011, pp. 245--301.

\bibitem{crsp}
{The University of Chicago Booth School of Business}, ``Center for research in
  security prices {(CRSP)},'' 2015.

\bibitem{cmsdata}
\BIBentryALTinterwordspacing
CMS, ``Part {C} and {D} performance data,'' 2016. [Online]. Available:
  \url{https://www.cms.gov/Medicare/Prescription-Drug-Coverage/PrescriptionDrugCovGenIn/PerformanceData.html}
\BIBentrySTDinterwordspacing

\bibitem{gavrilov_2000}
M.~Gavrilov, D.~Anguelov, P.~Indyk, and R.~Motwani, ``Mining the stock market:
  Which measure is best?'' in \emph{Proceedings of the Sixth {ACM} {SIGKDD}
  International Conference on Knowledge Discovery and Data Mining}.\hskip 1em
  plus 0.5em minus 0.4em\relax New York, NY, USA: ACM, 2000, pp. 487--496.

\bibitem{cbo_2016}
``The {Budget} and {Economic} {Outlook}: 2016 to 2026,'' Jan. 2016,
  https://www.cbo.gov/publication/51129.

\bibitem{darden_2015}
M.~Darden and I.~M. McCarthy, ``The {Star} {Treatment}: {Estimating} the
  {Impact} of {Star} {Ratings} on {Medicare} {Advantage} {Enrollments},''
  \emph{Journal of Human Resources}, vol.~50, no.~4, pp. 980--1008, 2015.

\bibitem{reid_2013}
{Reid RO}, {Deb P}, {Howell BL}, and {Shrank WH}, ``Association between
  medicare advantage plan star ratings and enrollment,'' \emph{JAMA}, vol. 309,
  no.~3, pp. 267--274, Jan. 2013.

\bibitem{stuart_2011}
B.~Stuart, A.~Davidoff, R.~Lopert, T.~Shaffer, J.~Samantha~Shoemaker, and
  J.~Lloyd, ``\BIBforeignlanguage{en}{Does {Medication} {Adherence} {Lower}
  {Medicare} {Spending} among {Beneficiaries} with {Diabetes}?}''
  \emph{\BIBforeignlanguage{en}{Health Services Research}}, vol.~46, no.~4, pp.
  1180--1199, Aug. 2011.

\bibitem{roebuck_2011}
M.~C. Roebuck, J.~N. Liberman, M.~Gemmill-Toyama, and T.~A. Brennan,
  ``\BIBforeignlanguage{en}{Medication {Adherence} {Leads} {To} {Lower}
  {Health} {Care} {Use} {And} {Costs} {Despite} {Increased} {Drug}
  {Spending}},'' \emph{\BIBforeignlanguage{en}{Health Affairs}}, vol.~30,
  no.~1, pp. 91--99, Jan. 2011.

\bibitem{pittman_2011}
D.~G. Pittman, W.~Chen, S.~J. Bowlin, and J.~M. Foody, ``Adherence to
  {Statins}, {Subsequent} {Healthcare} {Costs}, and {Cardiovascular}
  {Hospitalizations},'' \emph{The American Journal of Cardiology}, vol. 107,
  no.~11, pp. 1662--1666, Jun. 2011.

\bibitem{pittman_2010}
D.~G. Pittman, Z.~Tao, W.~Chen, and G.~D. Stettin,
  ``\BIBforeignlanguage{eng}{Antihypertensive medication adherence and
  subsequent healthcare utilization and costs},''
  \emph{\BIBforeignlanguage{eng}{The American Journal of Managed Care}},
  vol.~16, no.~8, pp. 568--576, Aug. 2010.

\bibitem{soria-saucedo_2016}
R.~Soria-Saucedo, P.~Xu, J.~Newsom, H.~Cabral, and L.~E. Kazis,
  ``\BIBforeignlanguage{eng}{The {Role} of {Geography} in the {Assessment} of
  {Quality}: {Evidence} from the {Medicare} {Advantage} {Program}},''
  \emph{\BIBforeignlanguage{eng}{PloS One}}, vol.~11, no.~1, p. e0145656, 2016.

\bibitem{couto_2014}
J.~E. Couto, J.~M. Panchal, L.~S. Lal, T.~J. Bunz, J.~E. Maesner, T.~O'Brien,
  and T.~Khan, ``\BIBforeignlanguage{eng}{Geographic variation in medication
  adherence in commercial and {Medicare} part {D} populations},''
  \emph{\BIBforeignlanguage{eng}{Journal of Managed Care \& Specialty
  Pharmacy}}, vol.~20, no.~8, pp. 834--842, Aug. 2014.

\end{thebibliography}

\begin{thebibliography}{10}
\providecommand{\url}[1]{#1}
\csname url@samestyle\endcsname
\providecommand{\newblock}{\relax}
\providecommand{\bibinfo}[2]{#2}
\providecommand{\BIBentrySTDinterwordspacing}{\spaceskip=0pt\relax}
\providecommand{\BIBentryALTinterwordstretchfactor}{4}
\providecommand{\BIBentryALTinterwordspacing}{\spaceskip=\fontdimen2\font plus
\BIBentryALTinterwordstretchfactor\fontdimen3\font minus
  \fontdimen4\font\relax}
\providecommand{\BIBforeignlanguage}[2]{{%
\expandafter\ifx\csname l@#1\endcsname\relax
\typeout{** WARNING: IEEEtran.bst: No hyphenation pattern has been}%
\typeout{** loaded for the language `#1'. Using the pattern for}%
\typeout{** the default language instead.}%
\else
\language=\csname l@#1\endcsname
\fi
#2}}
\providecommand{\BIBdecl}{\relax}
\BIBdecl

\bibitem{gavrilov_2000}
M.~Gavrilov, D.~Anguelov, P.~Indyk, and R.~Motwani, ``Mining the stock market:
  Which measure is best?'' in \emph{Proceedings of the Sixth {ACM} {SIGKDD}
  International Conference on Knowledge Discovery and Data Mining}.\hskip 1em
  plus 0.5em minus 0.4em\relax New York, NY, USA: ACM, 2000, pp. 487--496.

\bibitem{xu_2013}
K.~S. Xu, M.~Kliger, and A.~O. Hero~III, ``Adaptive evolutionary clustering,''
  \emph{Data Mining and Knowledge Discovery}, vol.~28, no.~2, pp. 304--336,
  Jan. 2013.

\bibitem{cbo_2016}
``The {Budget} and {Economic} {Outlook}: 2016 to 2026,'' Jan. 2016,
  https://www.cbo.gov/publication/51129.

\bibitem{darden_2015}
M.~Darden and I.~M. McCarthy, ``The {Star} {Treatment}: {Estimating} the
  {Impact} of {Star} {Ratings} on {Medicare} {Advantage} {Enrollments},''
  \emph{Journal of Human Resources}, vol.~50, no.~4, pp. 980--1008, 2015.

\bibitem{reid_2013}
{Reid RO}, {Deb P}, {Howell BL}, and {Shrank WH}, ``Association between
  medicare advantage plan star ratings and enrollment,'' \emph{JAMA}, vol. 309,
  no.~3, pp. 267--274, Jan. 2013.

\bibitem{stuart_2011}
B.~Stuart, A.~Davidoff, R.~Lopert, T.~Shaffer, J.~Samantha~Shoemaker, and
  J.~Lloyd, ``\BIBforeignlanguage{en}{Does {Medication} {Adherence} {Lower}
  {Medicare} {Spending} among {Beneficiaries} with {Diabetes}?}''
  \emph{\BIBforeignlanguage{en}{Health Services Research}}, vol.~46, no.~4, pp.
  1180--1199, Aug. 2011.

\bibitem{roebuck_2011}
M.~C. Roebuck, J.~N. Liberman, M.~Gemmill-Toyama, and T.~A. Brennan,
  ``\BIBforeignlanguage{en}{Medication {Adherence} {Leads} {To} {Lower}
  {Health} {Care} {Use} {And} {Costs} {Despite} {Increased} {Drug}
  {Spending}},'' \emph{\BIBforeignlanguage{en}{Health Affairs}}, vol.~30,
  no.~1, pp. 91--99, Jan. 2011.

\bibitem{pittman_2011}
D.~G. Pittman, W.~Chen, S.~J. Bowlin, and J.~M. Foody, ``Adherence to
  {Statins}, {Subsequent} {Healthcare} {Costs}, and {Cardiovascular}
  {Hospitalizations},'' \emph{The American Journal of Cardiology}, vol. 107,
  no.~11, pp. 1662--1666, Jun. 2011.

\bibitem{pittman_2010}
D.~G. Pittman, Z.~Tao, W.~Chen, and G.~D. Stettin,
  ``\BIBforeignlanguage{eng}{Antihypertensive medication adherence and
  subsequent healthcare utilization and costs},''
  \emph{\BIBforeignlanguage{eng}{The American Journal of Managed Care}},
  vol.~16, no.~8, pp. 568--576, Aug. 2010.

\bibitem{cmsdata}
\BIBentryALTinterwordspacing
CMS, ``Part {C} and {D} performance data,'' 2016. [Online]. Available:
  \url{https://www.cms.gov/Medicare/Prescription-Drug-Coverage/PrescriptionDrugCovGenIn/PerformanceData.html}
\BIBentrySTDinterwordspacing

\bibitem{soria-saucedo_2016}
R.~Soria-Saucedo, P.~Xu, J.~Newsom, H.~Cabral, and L.~E. Kazis,
  ``\BIBforeignlanguage{eng}{The {Role} of {Geography} in the {Assessment} of
  {Quality}: {Evidence} from the {Medicare} {Advantage} {Program}},''
  \emph{\BIBforeignlanguage{eng}{PloS One}}, vol.~11, no.~1, p. e0145656, 2016.

\bibitem{couto_2014}
J.~E. Couto, J.~M. Panchal, L.~S. Lal, T.~J. Bunz, J.~E. Maesner, T.~O'Brien,
  and T.~Khan, ``\BIBforeignlanguage{eng}{Geographic variation in medication
  adherence in commercial and {Medicare} part {D} populations},''
  \emph{\BIBforeignlanguage{eng}{Journal of Managed Care \& Specialty
  Pharmacy}}, vol.~20, no.~8, pp. 834--842, Aug. 2014.

\end{thebibliography}

\end{document}